\documentclass[twoside,journal]{IEEEtran}
\usepackage{bm,amsmath,amsfonts}
\usepackage{algorithmic}
\usepackage{algorithm}
\usepackage{array}
\usepackage[caption=false,font=normalsize,labelfont=sf,textfont=sf]{subfig}
\usepackage{textcomp}
\usepackage{stfloats}
\usepackage{graphicx}
\usepackage{url}
\usepackage{threeparttable}
\usepackage{multirow}
\usepackage{wrapfig}
\usepackage{verbatim}
\usepackage{graphicx}

\usepackage{cite}
\usepackage{tabularx}
\usepackage[colorlinks,
linkcolor=blue,
anchorcolor=blue,
urlcolor = blue,
citecolor=blue]{hyperref}
  
\begin{document}
\title{Neural Combinatorial Optimization Algorithms for Solving Vehicle Routing Problems:     

A Comprehensive Survey with Perspectives}

\author{

Xuan Wu, Di Wang,~\IEEEmembership{Senior Member,~IEEE,} Lijie~Wen, Yubin~Xiao$^\star$, Chunguo~Wu, Yuesong~Wu, Chaoyu~Yu, Douglas~L.~Maskell,~\IEEEmembership{Senior~Member,~IEEE,} You~Zhou$^\star$

\thanks{
\emph{($^\star$ Corresponding authors: Yubin Xiao and You Zhou).}


Xuan Wu, Yubin Xiao, Chunguo Wu, Yuesong Wu, Chaoyu Yu, You Zhou are with the Key Laboratory of Symbolic Computation and Knowledge Engineering of Ministry of Education, College of Computer Science and Technology, Jilin university, Changchun 130012, China (e-mail: \{wuuu22, xiaoyb21, wuys23, zhaoyuy22\}@mails.jlu.edu.cn, \{wucg, zyou\}@jlu.edu.cn).

Di Wang is with the Joint NTU-UBC Research Centre of Excellence in Active Living for the Elderly, Nanyang Technological University, 639798, Singapore (e-mail: wangdi@ntu.edu.sg).

Lijie Wen is with the School of Software, Tsinghua University, Beijing 100084, China (e-mail: wenlj@tsinghua.edu.cn).

Douglas L. Maskell is with the College of Computing and Data Science, Nanyang Technological University, 639798, Singapore (e-mail: asdouglas@ntu.edu.sg).
}}

\maketitle
\begin{abstract}
Although several surveys on Neural Combinatorial Optimization (NCO) solvers specifically designed to solve Vehicle Routing Problems (VRPs) have been conducted, they did not cover the state-of-the-art (SOTA) NCO solvers emerged recently. More importantly, to establish a comprehensive and up-to-date taxonomy of NCO solvers, we systematically review relevant publications and preprints, categorizing them into four distinct types, namely Learning to Construct, Learning to Improve, Learning to Predict-Once, and Learning to Predict-Multiplicity solvers. Subsequently, we present the inadequacies of the SOTA solvers, including poor generalization, incapability to solve large-scale VRPs, inability to address most types of VRP variants simultaneously, and difficulty in comparing these NCO solvers with the conventional Operations Research algorithms. Simultaneously, we discuss on-going efforts, identify open inadequacies, as well as propose promising and viable directions to overcome these inadequacies. Notably, existing efforts focus on only one or two of these inadequacies, with none attempting to address all of them concurrently. In addition, we compare the performance of representative NCO solvers from the Reinforcement, Supervised, and Unsupervised Learning paradigms across VRPs of varying scales. Finally, following the proposed taxonomy, we provide an accompanying web page as a live repository for NCO solvers. Through this survey and the live repository, we aim to foster further advancements in the NCO community.
\end{abstract}

\section{Introduction}
\label{sec1}
\IEEEPARstart{C}{ombinatorial} Optimization Problem (COP) is a vital branch of mathematical optimization, dedicated to seeking optimal solutions for problems within the discrete space \cite{schrijver_history_2005, yang_goods_2020, wu_incorporating_2023}. Among various COPs, Vehicle Routing Problems (VRPs) have extensive practical applications across diverse domains, e.g., communication and transportation \cite{ ge_route_2021, xiao_distilling_2024, Zhou_Learning_2023, Kong_2024}. Generally speaking, Operations Research (OR) algorithms for solving VRPs can be broadly divided into three categories, namely exact, approximation, and heuristic algorithms \cite{li_research_2021}. Specifically, exact algorithms typically employ the divide-and-conquer (D\&C) manner to solve VRPs and obtain optimal solutions. These algorithms mainly encompass Branch and Bound \cite{lawler1966branch} and Dynamic Programming (DP) \cite{Bellman_1952_Theory}. Approximation algorithms, aimed at securing solutions of assured quality, include Relaxation Algorithm \cite{TOTH2002487} and Linear Programming \cite{Hochba_1997_28}. Heuristic algorithms represent a class of methods that utilize the predefined heuristic rules to efficiently explore solutions in a given timeframe, developed through continuous trial-and-error. The most prominent heuristic algorithm for solving VRPs is Lin-Kernighan-Helsgaun 3 (LKH3) \cite{helsgaun2017extension}. Additionally, it is noteworthy that approximation and heuristic algorithms may have the potential to converge to the optimal solutions given sufficient time, yet such convergence is not guaranteed \cite{wu_neural_2023, li_research_2021}.

\begin{figure}[!t]
\centering
	\includegraphics[scale=0.3]{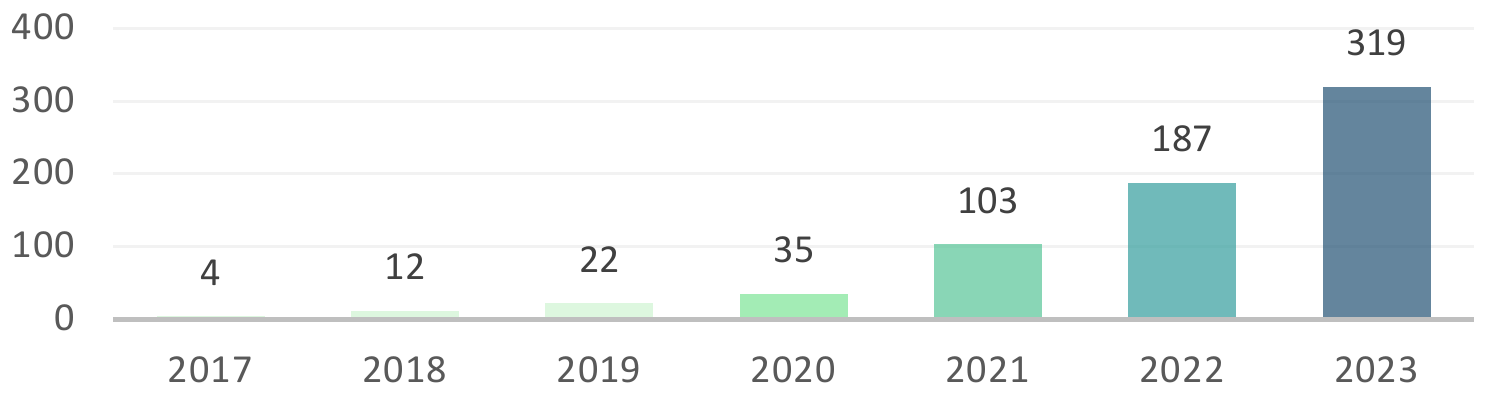}
	\caption{Illustration of the number of publications and preprints on NCO solvers for VRPs. This information is gathered from Google Scholar and Web of Science with the keywords ``Neural Combinatorial Optimization" \textit{OR} ``NCO" \textit{OR} ``Reinforcement Learning" \textit{OR} ``Deep Learning" \textit{OR} ``Neural Network" \textit{AND} ``Vehicle Routing Problem" \textit{OR} ``VRP" \textit{OR} ``Traveling Salesman Problem" \textit{OR} ``TSP" by the end of 2023. Following the initial data collection, a meticulous examination of each literature piece is conducted to precisely define its scope within the realm of NCO.} 
	\label{fig1}
\end{figure}

However, these three types of conventional OR algorithms lack the ability to derive insights from historical VRP instances, leading to significant computing overhead, especially for medium- or large-scale VRPs \cite{boussaid_survey_2013,sui_survey_2024}. This is demonstrated by the performance of LKH3 algorithm in solving a 100-node Capacitied Vehicle Routing Problem (CVRP) instance, requiring a noteworthy 12 hours to attain the optimal solution \cite{kwon_pomo_2020}. This shortcoming constrains the applications of OR algorithms in time-sensitive scenarios, such as on-call routing \cite{ghiani_real-time_2003} and ride-hailing service \cite{Xu_large_2018}. To alleviate the expensive computing overhead, many Neural Combinatorial Optimization (NCO) solvers utilizing Deep Learning (DL) have recently emerged \cite{bengio_machine_2021, liu_how_2023, li_deep_2023, grinsztajn2023winner, Wang_postive_2023, Liang_Splitnet_2023,xiao2024nips}, as shown in Fig.~\ref{fig1}. These NCO solvers possess the ability to learn from extensive historical instances and efficiently search for (sub-)optimal solutions, when dealing with instances exhibiting the same distribution characteristics. Moreover, by leveraging GPUs, these NCO solvers demonstrate efficiency in concurrently handling batches of instances, thereby further enhancing the speed of problem-solving \cite{zhang_review_2023, chalumeau2023combinatorial, vinyals_pointer_2015, kool_attention_2019}. For example, for a 100-node CVRP instance, an NCO solver named POMO \cite{kwon_pomo_2020} requires only one minute to find the sub-optimal solution with a gap value of less than 1\%. Here, the gap denotes the relative difference in the ``average length” between POMO and LKH3. 

In the past three years, a number of surveys \cite{zhang_review_2023, bogyrbayeva_learning_2022, li_research_2021, bengio_machine_2021, cappart_combinatorial_2021,  mazyavkina_reinforcement_2021, parvez_farazi_deep_2021, wang_deep_2021, li_overview_2022, liu_how_2023, SHI2023773} have been conducted to introduce and summarize NCO solvers specifically designed for VRPs. Nonetheless, these existing surveys have various limitations. First, the taxonomies proposed in these surveys \cite{zhang_review_2023, bogyrbayeva_learning_2022, li_research_2021, bengio_machine_2021, cappart_combinatorial_2021} do not facilitate an appropriate category for all NCO solvers. Secondly, a number of surveys \cite{li_research_2021, bengio_machine_2021, cappart_combinatorial_2021, mazyavkina_reinforcement_2021, parvez_farazi_deep_2021, wang_deep_2021, liu_how_2023, SHI2023773, li_overview_2022} do not adequately cover current NCO solvers' inadequacies and associated efforts. Thirdly, a few surveys \cite{mazyavkina_reinforcement_2021, parvez_farazi_deep_2021, wang_deep_2021} exclusively focus on NCO solvers utilizing the Reinforcement Learning (RL) paradigm, overlooking those based on Supervised Learning (SL) and Unsupervised Learning (UL). Consequently, these surveys do not provide a comprehensive overview of state-of-the-art (SOTA) NCO solvers. Finally, many surveys \cite{zhang_review_2023, li_research_2021, bengio_machine_2021, cappart_combinatorial_2021, mazyavkina_reinforcement_2021, parvez_farazi_deep_2021} focus on multiple COPs, without providing a detailed introduction to VRPs. All these necessitate a more up-to-date survey to include the SOTA advancements in this important research field. In the following paragraphs, we elaborate on these four limitations of existing surveys and introduce the advantages of this survey.

Regarding the first limitation, the two taxonomies proposed in prior surveys \cite{zhang_review_2023, bogyrbayeva_learning_2022, li_research_2021, bengio_machine_2021, cappart_combinatorial_2021} fall short in appropriately classifying all NCO solvers. Specifically, based on the adopted learning paradigm, the first taxonomy \cite{zhang_review_2023} classifies NCO solvers into three categories, namely solvers adopting Constructive Deep RL, Improving Deep RL, and SL or RL. However, the solving process of SL solvers  (e.g.,\cite{vinyals_pointer_2015, yao2023dataefficient, luo_neural_2023}) aligns with solvers adopting Constructive Deep RL (e.g., \cite{kool_attention_2019, kwon_pomo_2020}). In essence, either type of solvers \cite{vinyals_pointer_2015, yao2023dataefficient, luo_neural_2023} or \cite{kool_attention_2019, kwon_pomo_2020} employ the learned heuristics to select the unvisited nodes and add them to partial solutions sequentially. Consequently, it is not appropriate to classify NCO solvers solely based on the adopted learning paradigm. The other taxonomy categorizes NCO solvers into three types, namely End to End (EE), Learning to Configure (LC), and Multiple Calls (MC) solvers \cite{bengio_machine_2021}. EE solvers employ a Neural Network (NN) dedicated to solving VRPs in isolation, LC solvers offer the parameters of OR algorithms (only once), and MC solvers assist OR algorithms in making lower-level decisions multiple times throughout the search process. However, certain NCO solvers do not fit into any of the categories in the second taxonomy. For example, Joshi et al. \cite{joshi_efficient_2019} predicted key information only once in the whole search process of Beam Search, diverging from providing parameters for OR algorithms, thereby placing this solver outside the categories proposed in \cite{bengio_machine_2021}. Therefore, the second taxonomy has its drawbacks in categorizing all NCO solvers.

\begin{table}[!t]
\centering
\caption{Our proposed taxonomy of NCO solvers}
				\label{t2}
				\scalebox{0.8}{
				\begin{tabularx}{\columnwidth}{|>{\centering\arraybackslash}m{0.1\columnwidth}|>{\centering\arraybackslash}m{0.36\columnwidth}|>{\centering\arraybackslash}m{0.39\columnwidth}|}

			\hline
			Category & Description & Representatives\\ \hline
			L2C  &  Using NNs to construct solutions from scratch & Ptr-Net \cite{vinyals_pointer_2015}; AM \cite{kool_attention_2019};  POMO \cite{kwon_pomo_2020}; NAR4TSP\cite{xiao2023reinforcement} \\ \hline
			L2I  &  Using NNs to iteratively improve complete solutions & NeuRewriter \cite{chen_learning_2019}; MT \cite{wu_learning_2022}; N2OPT \cite{da_costa_learning_2020}; NeuOPT \cite{ma_learning_2023}  \\ \hline
			L2P-O & Using NNs to assist OR algorithms once & GCN \cite{joshi_efficient_2019};  NeuroLKH \cite{xin_neurolkh_2021};  DPDP \cite{kool_deep_2022}; DIFUSCO \cite{sun_difusco_2023} \\ \hline
			L2P-M & Using NNs to assist OR algorithms multiplicity & GAVE \cite{hottung_learning_2021}; VSR-LKH  \cite{zheng_combining_2021};  MOCO \cite{dernedde2024moco}; EOH \cite{liu2024example}  \\ \hline
		\end{tabularx}}
\end{table}

In this survey, we propose a novel taxonomy aimed at appropriately and exclusively classifying all NCO solvers. As shown in Table~\ref{t2}, we categorize these solvers into four categories, namely Learning to Construct (L2C), Learning to Improve (L2I), Learning to Predict-Once \mbox{(L2P-O)}, and Learning to Predict-Multiplicity (L2P-M) solvers. L2C solvers represent the pioneering category of NCO solvers, drawing inspiration from the analogy between VRPs and the machine translation task. Specifically, L2C solvers use NNs to sequentially select nodes that yet to be visited (resembling the process of translating words in the machine translation task) and add them to partial solutions \cite{kool_attention_2019, bresson_transformer_2021, xin_multi-decoder_2021}. For example, Vinyals et al. \cite{vinyals_pointer_2015} were the pioneers who employed SL to train a Recurrent Neural Network (RNN) for sequentially constructing solutions for Traveling Salesman Problem (TSP). Subsequently, inspired by the notion of iteratively ruining and repairing in heuristic algorithms \cite{SCHRIMPF2000139}, certain L2I solvers \cite{Zhou_Learning_2023, chen_learning_2019, hottung_neural_2022, kim2023learning} introduce NNs to emulate this process, aiming to derive the (sub-)optimal solutions. For example, Chen and Tian \cite{chen_learning_2019} selectively deconstructed the local components of complete solutions and repaired them through a learned policy, enhancing the quality of current complete solutions. Meanwhile, certain \mbox{L2P-O} solvers \cite{joshi_efficient_2019, kool_deep_2022} and L2P-M solvers \cite{zheng_combining_2021, hottung_learning_2021} diverge from emulating the iteratively improving process. Instead, these solvers focus on integrating the predicted key information with OR algorithms to enhance solution quality and/or reduce search overhead. Furthermore, L2P-O solvers predict information only once throughout the entire search process of OR algorithms, whereas L2P-M ones require the facilitation of OR algorithms during the whole search process, making predictions and decisions multiple times. For example, to expedite the search process of DP, Kool et al. \cite{kool_deep_2022} proposed an \mbox{L2P-O} solver named Deep Policy Dynamic Programming (DPDP). This solver predicts promising edges for each TSP instance before the initiation of the DP search. The solution is then exclusively constructed using these identified promising edges during the subsequent DP search. In line with the objective of \cite{kool_deep_2022} to expedite the search process of OR algorithms, an L2P-M solver named VSR-LKH \cite{zheng_combining_2021} exploits RL to inform decisions at each search step of the LKH algorithm \cite{helsgaun_effective_2000}. In conclusion, the taxonomy proposed in this paper provides a structured framework to comprehensively present the diversity of NCO solvers, ensuring that all solvers can be exclusively categorized into one of the four categories defined in our taxonomy. In Sections~\ref{sec3} to \ref{sec5}, we discuss the reasonableness and exclusivity of the proposed taxonomy, providing comprehensive overviews of the relevant studies corresponding to each solver category.

The second limitation arises from the lack of comprehensive reporting on the inadequacies of SOTA NCO solvers in prior surveys \cite{li_research_2021, bengio_machine_2021, cappart_combinatorial_2021, mazyavkina_reinforcement_2021, parvez_farazi_deep_2021, wang_deep_2021, liu_how_2023}. Unfortunately, they only introduce the relatively outdated inadequacies of NCO solvers, failing to provide researchers and practitioners with insights into the latest developments in NCO solvers. For example, certain surveys identify the performance of NCO solvers on small- or medium-scale VRPs as a potential inadequacy. However, recent NCO solvers (e.g., GELD \cite{xiao2025}) have effectively addressed this inadequacy, achieving performance on par with exact algorithms. Again, a more up-to-date survey is required to reflect these recent advancements. To identify the open inadequacies of SOTA NCO solvers, we conduct a systematic survey and simultaneously introduce on-going efforts to address four open inadequacies. Specifically, the first inadequacy is that existing NCO solvers are only capable of solving VRPs that have the same instance distribution \cite{joshi_learning_2021, bi_learning_2022, jiang_learning_2022, fang2024invit}. For example, NCO solvers trained on a type of specific instance distribution (e.g., the uniform distribution) might not perform well when being applied onto instances from other distribution patterns (e.g., the cluster distribution) \cite{bi_learning_2022}. In addition, existing NCO solvers face inadequacies in addressing large-scale VRPs (e.g., the 10,000-node TSP) in real-time \cite{cheng_select_2023, pan_h-tsp_2023, liu2023edgeaware}. Moreover, existing NCO solvers exhibit limited performance on various VRP variants (e.g., Asymmetric Traveling Salesman Problem \cite{kwon_matrix_2021}), necessitating the development of specific NNs for each VRP variant. Finally, the issue of unfair comparisons emerges as a substantial inadequacy in evaluating NCO solvers against the conventional OR algorithms, resulting in a constrained acceptance of these NCO solvers within the OR community  \cite{garmendia2022neural, accorsi_guidelines_2022}. To better solve these inadequacies, recent studies \cite{xiao2025,mvmoe,zheng2024udc} have explored various approaches and proposed corresponding solvers, which have been reviewed in this survey. More importantly, apart from introducing on-going efforts to address these inadequacies, we also identify open inadequacies and suggest promising research directions to address them (see Section~\ref{sec6} for more details). It is worth highlighting that while certain prior studies (e.g., GELD \cite{xiao2025}) have attempted to tackle one or two of these inadequacies, none simultaneously addresses three or even all four inadequacies, and the overall performance of many NCO solvers still leaves room for improvement (see Section~\ref{sec6.3} for more details).

The third limitation stems from the fact that certain surveys \cite{mazyavkina_reinforcement_2021, parvez_farazi_deep_2021, wang_deep_2021, panzer_deep_2022} exclusively focus on NCO solvers utilizing RL, overlooking those based on SL and UL. Consequently, these surveys do not provide a comprehensive performance comparison across all solvers, which is a non-negligible issue due to the fact that certain SOTA SL- and UL-based solvers \cite{min2023unsupervised, sun_difusco_2023} have demonstrated high-level performance on large-scale VRPs. In contrast, our survey stands out by introducing NCO solvers based on all learning paradigms, namely SL, UL, and RL. For example, we introduce the SL-based solver LEHD \cite{luo_neural_2023} and the UL-based solver UTSP \cite{min2023unsupervised}, both of which were specifically designed to solve large-scale VRPs by reducing the number of parameters (see Section~\ref{sec6.2} for more details). By employing this comprehensive survey methodology, this survey paper provides a thorough performance comparison among solvers across varying scales of VRPs.

Finally, a number of existing surveys \cite{zhang_review_2023, li_research_2021, bengio_machine_2021, cappart_combinatorial_2021, mazyavkina_reinforcement_2021, parvez_farazi_deep_2021} have delved into a spectrum of COPs, such as Job Shop Scheduling Problem (JSP) \cite{zhang_learning_2020, Ni_multi_2021, zhang2024deep,zhao2025dual} and Knapsack Problem (KP) \cite{zhao2022learning, Jiang_Learning_2023, cheng2024learning}. Consequently, these surveys are unable to provide a comprehensive and detailed introduction to VRPs. In contrast, this survey paper takes a focused approach to comprehensively introduce the details of NCO solvers tailored for VRPs, the most researched COPs in the NCO domain \cite{min2024size}. For example, in Section~\ref{sec3.4}, we elaborate on data augmentation methods specially designed for VRPs, which are often overlooked in other surveys. Moreover, another reason for the exclusive focus of this survey on VRPs is that most NCO studies initially focused on VRPs before extending their applications to other COPs. For example, Kool et al. \cite{kool_attention_2019} pioneered the integration of Attention Model (AM) into NCO to solve VRPs, which is expanded upon by subsequent studies \cite{zhang2024deep} and \cite{Jiang_Learning_2023} to tackle JSPs and KPs, respectively. Hence, it is justified for us to purposely focus on VRPs in this survey.

The key contributions of this work are as follows:
\begin{itemize} 

\item This survey paper proposes a novel taxonomy for NCO solvers specifically designed to solve VRPs. Specifically, the proposed taxonomy categorizes existing NCO solvers into four exclusive types. 

\item This paper outlines the inadequacies of existing NCO solvers, including generalization, large-scale and multi-constrained VRPs, and fair comparisons with OR algorithms, while introducing on-going efforts and suggesting promising directions to better address these inadequacies.

\item This paper explores NCO solvers across diverse learning paradigms for VRPs, introduces them in a detailed manner, and comprehensively assesses their performance across different problem scales. 

\item This paper provides an accompanying web page\footnote{URL: https://github.com/wuuu110/NCO-solvers-for-VRP} as a live repository of papers on NCO solvers organized according to our proposed taxonomy. In addition, this web page will be continuously updated with emerging studies.
\end{itemize}

\section{Preliminary}
\label{sec2}
Section~\ref{sec2.1} first introduces the definitions of TSP and CVRP. Subsequently, Section~\ref{sec2.2}  presents three learning paradigms of DL, namely SL, UL, and RL.

\subsection{Definitions of TSP and CVRP}
\label{sec2.1}
In this subsection, we specifically focus on two kinds of VRP variants, namely TSP and CVRP, due to their importance and prevalence \cite{kool_attention_2019}. For TSP, let $G\left(\bm{V}, \bm{E}\right)$ denote the graph of a VRP with $N$ cities, where $\bm{V}$ and $\bm{E}$ denote the vertex (i.e., $|\bm{V}| = N$) and edge sets, respectively. The edge $e_{i,j} \in \bm{E}$ denotes the Euclidean distance between the $i$th node $v_i$ and $j$th node $v_j$. In addition, $x_{i,j} =1$ indicates that $v_j$ is visited directly after $v_i$. Conversely, $x_{i,j} = 0$ indicates that $v_j$ is not visited directly after $v_i$. Finally, let $\tau_i$ denotes the visiting order of the $i$th node. Therefore, the Miller–Tucker–Zemlin formulation \cite{Miller_1960_integer} of TSP is given as follows:
\begin{equation}
\label{eq1}
\begin{array}{ll}
 \min\sum_{i=1}^N \sum_{j=1,j \neq i}^N e_{i,j}\cdot x_{i,j}, & x_{i,j} \in \{0,1\},
\end{array}
\end{equation}
\begin{equation}
\label{eq2}
 s.t. \begin{array}{ll} \sum_{i=1,i\neq j}^N x_{i,j}=1, & \forall j\in \{1,2,...,N\},  \end{array}
\end{equation}
\begin{equation}
\label{eq3}
 \begin{array}{ll} \sum_{j=1, j\neq i}^N x_{i,j}=1, & \forall i\in \{1,2,...,N\},  \end{array}
\end{equation}
\begin{equation}
\label{eq4}
 \begin{array}{ll} \tau_i-\tau_j+ N\cdot x_{i,j} \leq N-1, & 1 \leq i \neq j \leq N. \end{array}
\end{equation}
For TSP, the objective is to minimize the total distance cost of tour (solution) computed according to (\ref{eq1}). Constraints~(\ref{eq2}) and (\ref{eq3}) ensure that each node can be visited only once. Finally, constraint~(\ref{eq4}) eliminates sub-tours. 

Comparing to TSP, CVRP has the same objective, i.e., minimize the total distances of tour. But CVRP has more constraints due to varying goods demands in each customer and a maximum capacity limit for the delivery vehicle (assuming only one vehicle is available for all CVRP instances in this paper). Constrained by the capacity constraint, the vehicle visits partial nodes and subsequently returns to the depot; this sub-tour is termed a route. Specifically, let the $0$th node and $d_i$ denote the depot node and the demand of the $i$th customer node, respectively. In addition, let $e_{i,j}$, $u_i$, and $C$ denote the distance cost of a vehicle going from $i$th node to $j$th node, where $i,j \in \{0,1,2,...,N\}$, the surplus capacity of the vehicle after it serves the $i$th customer node, and the maximum capacity of a vehicle, respectively. An integer programming formulation of CVRP \cite{TOTH2002487} is given as follows:
\begin{equation}
\label{eq6}
 \begin{array}{ll} \min\sum_{i=0}^N \sum_{j=0}^N e_{i,j}\cdot x_{i,j}, & x_{i,j} \in \{0,1\}, \end{array}
\end{equation}
\begin{equation}
\label{eq7}
 s.t. \begin{array}{ll} \sum_{i=0, i\neq j}^N x_{i,j}=1, & \forall j\in \{1,2,...,N\}, \end{array}
\end{equation}
\begin{equation}
\label{eq8}
 \begin{array}{ll} \sum_{j=0, j\neq i}^N x_{i,j}=1, & \forall i\in \{1,2,...,N\},  \end{array}
\end{equation}
\begin{equation}
\label{eq9}
u_j=\left\{\begin{array}{ll}
		u_i-d_j, & x_{i,j} = 1, u_i > d_j, j\neq 0, \\
		C, & x_{i,j} = 1, j = 0, \\
			\end{array}\right.
\end{equation}
\begin{equation}
\label{eq10}
 \begin{array}{ll}  0 \leq d_i, u_i \leq C, & \forall i \in \{1,2,...,N\}. \end{array}
\end{equation}

For CVRP, constraints~(\ref{eq7}) and (\ref{eq8}) ensure that each node can be visited only once, apart from the depot node. Finally, constraints~(\ref{eq9}) and (\ref{eq10}) impose the vehicle capacity requirements.

\begin{table}[!t]
\centering
\caption{Different VRP variants and corresponding NCO solvers}
				\label{table1}
				\scalebox{0.8}{

		\begin{tabularx}{\columnwidth}{|>{\centering\arraybackslash}m{0.1\columnwidth}|m{0.46\columnwidth}|>{\centering\arraybackslash}m{0.29\columnwidth}|}
			\hline
			\multicolumn{1}{|c|}{Variant} & \multicolumn{1}{c|}{Description} & \multicolumn{1}{c|}{Corresponding Studies} \\ \hline
			TSP & All nodes are visited once, starting and ending with the same node. & \cite{vinyals_pointer_2015, kool_attention_2019, kwon_pomo_2020, ma_learning_2021, xin_multi-decoder_2021, cheng_select_2023, pan_h-tsp_2023, kim_sym-nco_2022, hou_generalize_2023, wu_learning_2022, bello_neural_2017, khalil_learning_2017, nazari_reinforcement_2018, drori2020learning, Xin_step_2021, kim_learning_2021, xu_reinforcement_2022, hottung_efficient_2022, drakulic2023bqnco, Son_meta_2023, Sun_Learning_2024} \\ \hline
			CVRP  & The cumulative demands of visited customer nodes for a single route must not surpass the assigned vehicle capacity. &\cite{kool_attention_2019, kwon_pomo_2020, ma_learning_2021, xin_multi-decoder_2021, chen_learning_2019, kim_sym-nco_2022, hou_generalize_2023, wu_learning_2022, bello_neural_2017, Xin_step_2021, kim_learning_2021, xu_reinforcement_2022, hottung_efficient_2022, hottung_neural_2020, lu_learning-based_2020, Zhao_hybird_2021, Son_meta_2023, Sun_Learning_2024} \\  \hline
			VRPTW & Node visits must adhere to specified time windows. Vehicles can wait in the event of early arrival, but late arrival is not permissible.  &  \cite{Zhao_hybird_2021, zhang_multi_2020, zhang_learning_2023, lin_deep_2022, chi_deep_2022, wu_solving_2021, ma_combinatorial_2019, falkner2020learning, silva_reinforcement_2019, Dang_warm_2023, chen2024looking,bi2024learning} \\  \hline
			MOVRP & In addition to minimizing tour length, additional objectives include minimizing traveling time, etc. & \cite{Wu_2020_MODRLDAM, Li_deep_2021, Zhang_meta_2022, zhang_MODRL_2021, Cai_Collaborative_2021, lin_pareto_2022, chen2023efficient, chen2023neural, Wang_Multiobjective_2023, luo24a,lu2025context,10478025}  \\ \hline
			DVRP  & Dynamically scheduling vehicles to achieve online-generated orders across multiple locations. &  \cite{Li_learning_2021b, ma2021a, zhang_solving_2023,  paul_multi_2022, sivagnanam_offline_2022, joe2020deep, Yu_online_2019, pan_deep_2023}  \\ \hline
			PDVRP & Visiting the pickup node prior to the associated delivery node in a service request. &  \cite{li_heterogeneous_2022, ma_efficient_2022, zong_mapdp_2022}\\ \hline
		\end{tabularx}}
\end{table}

The objectives of other VRP variants may be subject to different problem-specific constraints. For example, in the case of VRP with Time Window (VRPTW), node visits must adhere to specified time windows. In Table~\ref{table1}, apart from TSP and CVRP, we also present four widely studied VRP variants, namely VRPTW, Multi-Objective VRP (MOVRP), Dynamic VRP (DVRP), and Pickup-and-Delivery VRP (PDVRP). As indicated by the little overlap on the references listed in Table~\ref{table1}, most solvers cannot simultaneously address all or most types of VRP variants. Therefore, researchers must develop specialized NCO solvers to address VRP variants with additional constraints. See Section~\ref{sec6.3} for in-depth insights into the design principles of NCO solvers for solving VRPs with more constraints.

\subsection{Learning Paradigms of Deep Learning}
\label{sec2.2}
In this subsection, we introduce three kinds of DL training paradigms, namely SL, UL, and RL, where RL is widely adopted by NCO solvers for training NNs \cite{mazyavkina_reinforcement_2021, parvez_farazi_deep_2021, wang_deep_2021}.

\subsubsection{\textbf{Supervised Learning (SL)}}

SL is a paradigm in DL where the model undergoes training on a labeled dataset, containing input-output pairs. The objective of the model is to learn a function that produces predicted outputs closely matching the corresponding output labels, when given a set of input data. In the field of NCO, Vinyals et al. \cite{vinyals_pointer_2015} were the first to use a supervised loss function for training an RNN. The loss is computed as follows:
\begin{equation}
\label{eq12}
      \textit{loss} =-\sum\nolimits_{\mathcal{S},\tau^*}{\log{p_\theta(\tau^*|\mathcal{S})}},
\end{equation}
where $\theta$, $\mathcal{S}$, and $\tau^*$ denote the parameters of the RNN model, an instance with a number of $N$ nodes drawn from a specific type of distribution $\mathbb{S}$, and a permutation of integers representing the optimal tour of $\mathcal{S}$, respectively. Finally, $p_\theta(\tau^*|\mathcal{S})$ denotes the conditional probability of output $\tau^*$ given input $\mathcal{S}$, which is computed as follows: 
\begin{equation}
\label{eq11}
      p_\theta(\tau^*|\mathcal{S}) = \prod\nolimits_{i=1}^{N}{p(\tau^*_i|\tau^*_{(j, j<i)}, \mathcal{S};\theta)}.
\end{equation} 

However, acquiring labels for VRP instances, especially for large-scale VRPs, is often impractical or time-intensive. Furthermore, models trained using SL tend to show poor generalization, because they tend to capture noise or specific patterns that fail to generalize to new and unseen data \cite{bello_neural_2017}. These issues limit the widespread adoption of SL in solving VRPs \cite{kwon_pomo_2020}. Recently, to reduce the computing overhead of obtaining labels, certain studies \cite{yao2023dataefficient} proposed various data augmentation methods (see Section~\ref{sec3.4} for more details).

\subsubsection{\textbf{Reinforcement Learning (RL)}}

\begin{figure}[!t]
\centering
	\includegraphics[scale=0.5]{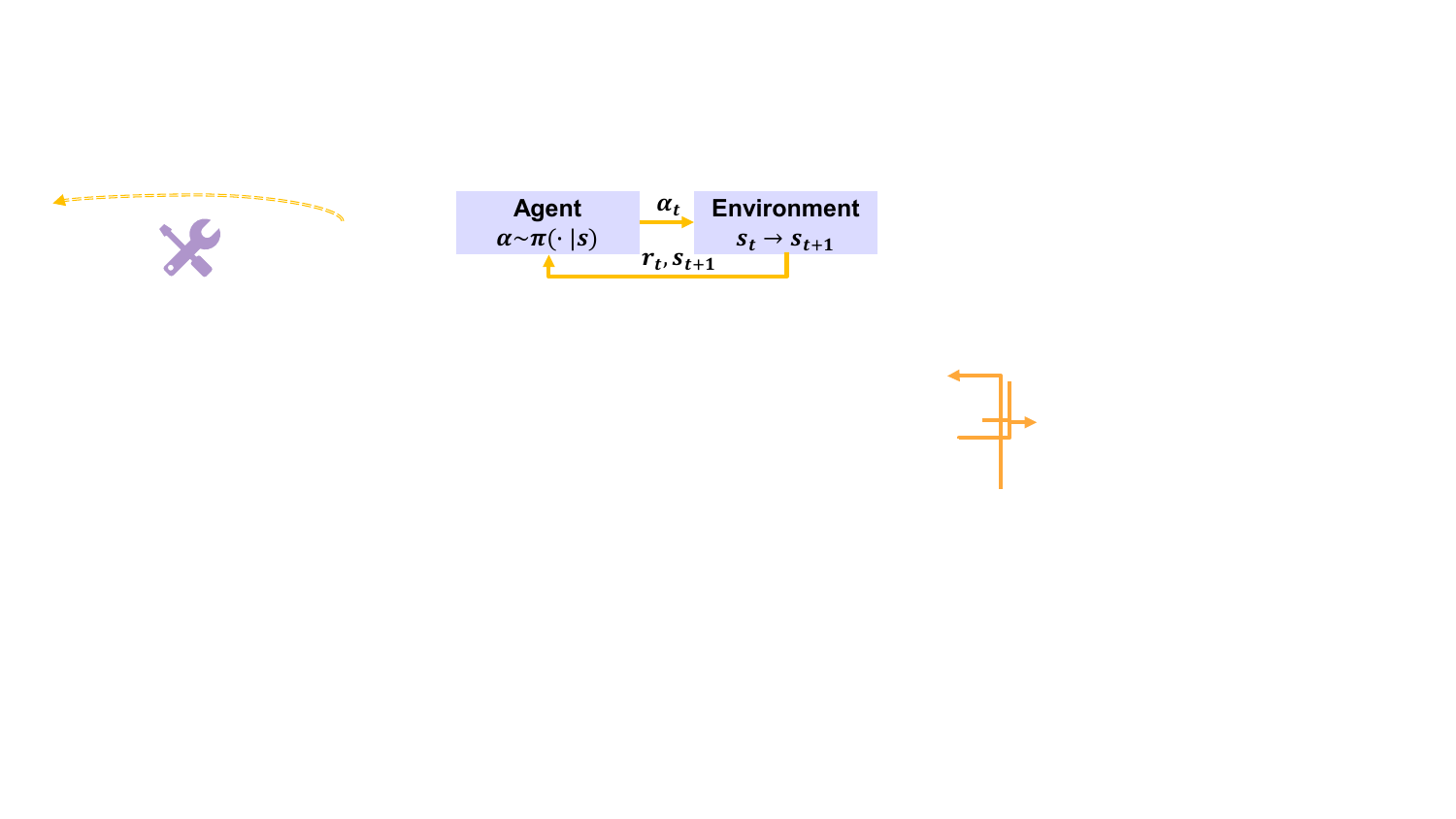}
	\caption{Illustration of interactions between the agent and the environment in RL, where $t$, $a_t$, $s_t$, $\pi$, and $r_t$ denote the time, action, state, policy, and reward, respectively. The policy $\pi(\cdot|s)$ of the agent takes the state $s_t$ as input and selects the action $a_t$ as output. Following this, the action is executed in the environment, and the agent receives a reward $r_t$. This sequential process continues as the state transitions to $s_{t+1}$.} 
	\label{fig3}
\end{figure}

RL represents a distinct paradigm in DL, where agent(s) acquires decision-making abilities through interactions with the environment. As shown in Fig.~\ref{fig3}, we describe the interacting process between a single agent and the environment of RL. At time $t$, the agent selects an action $a_t$ according to the state $s_t$. Subsequently, the state transitions to $s_{t+1}$, and the environment provides feedback in the form of a reward $r_t$ for the agent. 

Compared with SL-based solvers, RL ones autonomously learn to solve VRPs without human guidance and the effort of data labeling, thereby addressing the inherent limitation of SL's reliance on high-quality labels. In addition, RL-based solvers consistently surpass SL-based solvers in terms of generalization due to RL's interactive and exploratory nature \cite{joshi_learning_2019}. Hence, following the study of \cite{bello_neural_2017}, who introduced RL for training NNs to solve VRPs in 2017, a large number of RL-based solvers emerged \cite{kool_attention_2019, kwon_pomo_2020, zheng_reinforced_2023, kim_sym-nco_2022, xin_multi-decoder_2021, jin_pointerformer_2023,dpn,Son_equity_2024}. 

Specifically, the objective of the solver proposed in \cite{bello_neural_2017} is to maximize the expected return (total reward) $J$, which is given as follows:
\begin{equation}
\label{eq13}
     J(\theta|\mathcal{S})= -\mathbb{E}_{\tau \sim p_\theta (\cdot|\mathcal{S})}{L(\tau|\mathcal{S})},
\end{equation}
where $L(\cdot)$ produces the length of instance $\mathcal{S}$. The policy $p_\theta (\cdot|\mathcal{S})$ selects a solution $\tau$ given a VRP instance $\mathcal{S}$, which is factorized and parameterized by parameters $\theta$ as follows:
\begin{equation}
\label{eq14}
     p_\theta (\tau|\mathcal{S}) = \prod\nolimits_{i=1}^{N}{p(\tau_i|\tau_{(j, j<i)},\mathcal{S};\theta)},
\end{equation}
where $\tau_i$ and $\tau_{(j, j<i)}$ denote the current action at time $i$ and the previous actions, respectively. To optimize the parameters $\theta$, Bello et al. \cite{bello_neural_2017} employed the well-known REINFORCE algorithm \cite{williams_simple_1992}, which is a policy gradient method incorporating stochastic gradient descent to compute the gradient of (\ref{eq14}). The gradient is computed as follows:
\begin{equation}
\small
\label{eq15}
      \nabla_\theta J(\theta|\mathcal{S})= \mathbb{E}_{\tau \sim p_\theta (\cdot|\mathcal{S})}{[(L(\tau|\mathcal{S}) - b(\mathcal{S}))\nabla_\theta \log{p_\theta (\tau|\mathcal{S})}]},
\end{equation}
\normalsize
where $b(\mathcal{S})$ denotes the baseline function, which remains independent to the solution $\tau$ and serves to estimate the expected tour length, thereby reducing the variance of the gradients. Specifically, in \cite{bello_neural_2017}, an exponential moving average of returns obtained by the policy network over time is employed as the baseline function $b(\mathcal{S})$. 

It is worth mentioning that more effective baselines have been proposed in subsequent studies \cite{kwon_pomo_2020, kim_sym-nco_2022, da_costa_learning_2020, jin_pointerformer_2023}. One of the most straightforward yet most efficient baselines is the shared baseline proposed in POMO \cite{kwon_pomo_2020}. Specifically, in contrast to \cite{bello_neural_2017} where the NN consistently selects the fixed node as the starting node to generate a single solution for each instance, POMO mandates the generation of multiple solutions with diverse starting nodes for each instance. Subsequently, the averaged return of a set of solutions generated from different starting nodes is employed as the baseline.

\subsubsection{\textbf{Unsupervised Learning (UL)}}
Unlike the other two learning paradigms, UL does not require explicitly (i.e., labels in SL) or implicitly (i.e., rewards in RL) given information for learning. Instead, it endeavors to capture certain characteristics of the observed random variables. Until now, the combination of UL and VRP has received little attention \cite{zhang_review_2023, bengio_machine_2021}. Consequently, this survey refrains from an extensive exploration of UL, addressing it only when necessary. For instance, in Section~\ref{sec6.2}, we present an L2P-O solver named UTSP \cite{min2023unsupervised}, which incorporates a UL-based loss function to conserve memory and consequently address large-scale VRPs.

\section{Learning to Construct (L2C) Solvers}
\label{sec3}

\begin{figure}[!t]
\centering
	\includegraphics[scale=0.7]{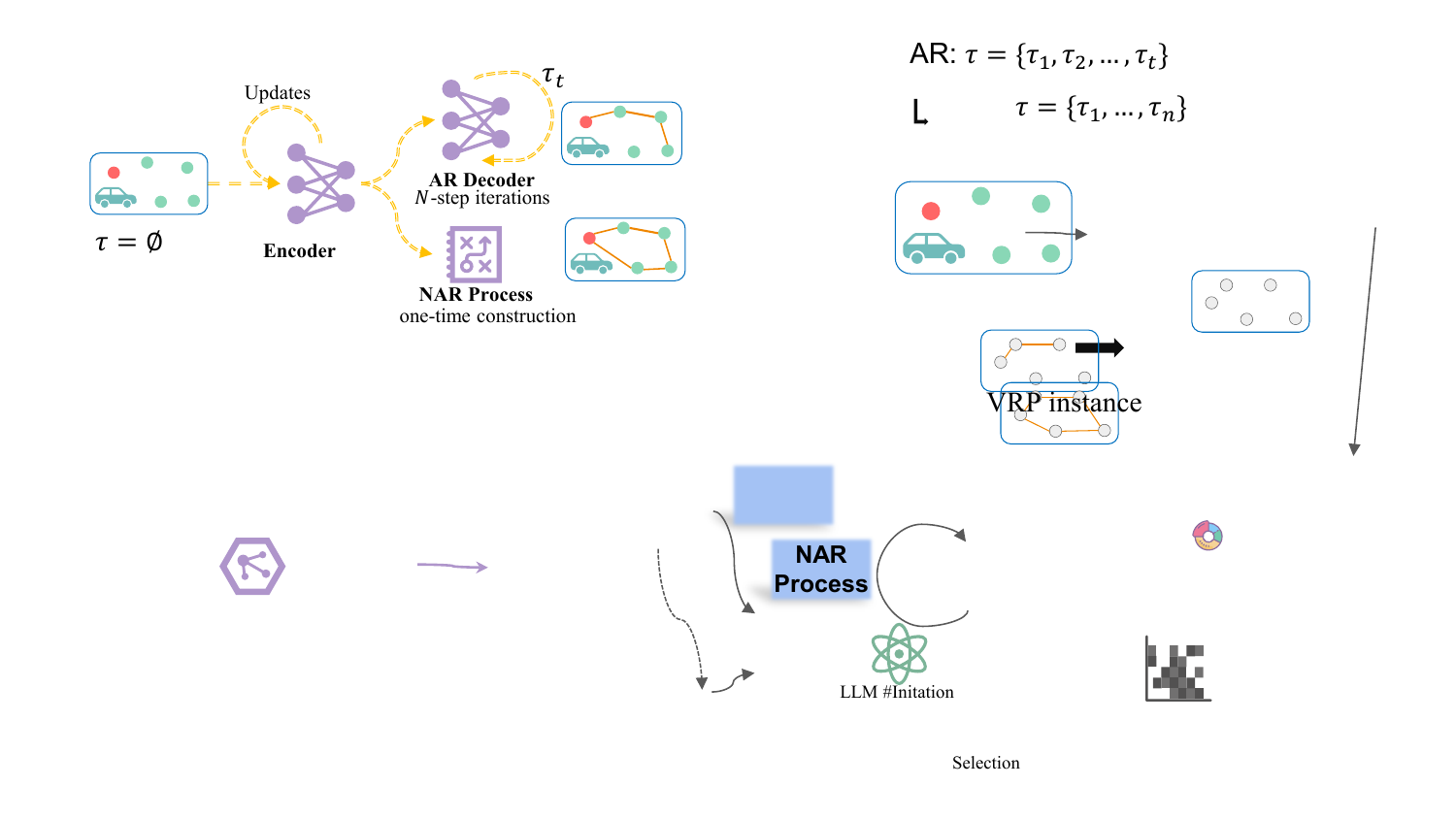}
	\caption{Illustration of the generic construction process of L2C solvers, starting from an empty solution set and ending with complete solutions. Most L2C solvers are composed of an encoder and a decoder. The encoder is used to output the embeddings of VRP instances, while the decoder selects nodes based on these embeddings to construct complete solutions. } 
	\label{fig4}
\end{figure}

As shown in Fig.~\ref{fig4}, similar to the machine translation task, L2C solvers start from an empty solution set, selecting unvisited nodes to construct complete solutions for VRPs. Generally speaking, L2C solvers are composed of an encoder and a decoder, predominantly leveraging a Markov Decision Process (MDP) formulation to learn how to solve VRPs. To offer a comprehensive review of L2C solvers, Sections~\ref{sec3.1} to \ref{sec3.3} introduce the design of the encoder, the design of the decoding process, and the MDP of L2C, respectively. Subsequently, Section~\ref{sec3.4} presents data augmentation methods and post-processing strategies. Finally, Section~\ref{sec3.5} compares the performance of L2C solvers on small-scale VRPs.

\subsection{Design of Encoders}
\label{sec3.1}
\begin{wrapfigure}[14]{r}{0.23\textwidth}
  \centering
  \includegraphics[width=0.23\textwidth]{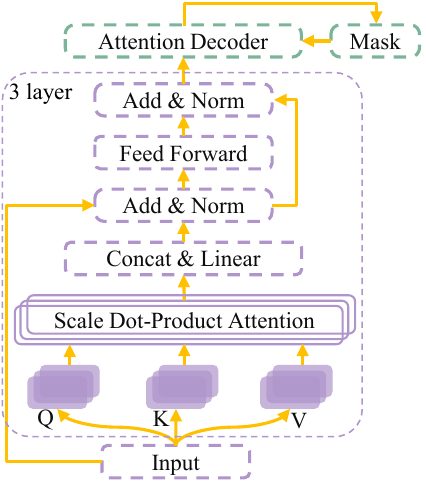}
  \caption{Illustration of AM \cite{kool_attention_2019}.}
  \label{figam}
\end{wrapfigure}

In the field of NCO, Bello et al. \cite{bello_neural_2017} were the first to employ actor-critic RL for training Long Short-Term Memory (LSTM) \cite{Hochreiter_1997_Long} as an encoder, capturing the coordinate information of nodes and converting it into embeddings. Subsequently, to fully leverage the graph structure, Khalil et al. \cite{khalil_learning_2017} proposed Graph Neural Network (GNN)-based S2V-DQN to embed node features. Differing from the previously employed RNN-based and GNN-based encoders, Kool et al. \cite{kool_attention_2019} proposed AM based on the superior Transformer architecture \cite{vaswani_attention_2017}. Specifically, as shown in Fig.~\ref{figam}, Kool et al. adopted a total of three attention layers as an encoder, with each layer comprising a multi-head attention and a node-wise feed-forward sub-layer. Ever since, AM has become a prevalent encoder model choice, adopted by many L2C solvers \cite{kwon_pomo_2020, kim_sym-nco_2022, xin_multi-decoder_2021, yuan_csl_2023, Son_equity_2024, hottung2024polynet} as the backbone structure. For example, to pursue better performance, Kwon et al. \cite{kwon_pomo_2020} further extended the number of attention layers of AM from three to six. Furthermore, the following paragraphs present two primary Transformer-based encoder designs tailored for specific problem scales and VRPs.

The first design involves certain L2C solvers \cite{luo_neural_2023, jin_pointerformer_2023, yang_memory_2023} propose innovative Transformer-based encoders with the objective of reducing the number of parameters to solve large-scale VRPs. This design choice stems from the observation that with the increasing scale of VRP, Transformer-based solvers experience a substantial surge in computation time and memory overhead. Consequently, L2C solvers endowed with a large number of parameters struggle to effectively address large-scale VRP instances \cite{luo_neural_2023}. To better solve this problem, Jin et al. \cite{jin_pointerformer_2023} adopted the Reversible Residual Network \cite{gomez2017reversible} instead of the standard Residual Network in the Transformer-based encoder. Reversible Residual Network does not require a large number of parameters by eliminating the need to store activation values in most layers. These values can be accurately reconstructed based on the activation values of the subsequent layer. Meanwhile, Yang et al. \cite{yang_memory_2023} replaced scaled dot-product attention with sampled scaled dot-production attention \cite{zhou2021informer}, where only the query and key of a small number of sampled nodes are computed rather than those of all nodes. While these solvers do reduce the number of parameters to some extent, their effectiveness in addressing large-scale VRPs still needs improvement, compared with other solvers (see Section~\ref{sec6.2} for more details).

Meanwhile, as shown in Fig.~\ref{fig4}, certain L2C solvers opt for the second design, i.e., real-time embedding updates. This design stems from the fact that many L2C solvers do not dynamically update node embeddings throughout the decoding process, potentially hindering their ability to model the dynamics of state transitions \cite{nazari_reinforcement_2018, xu_reinforcement_2022, peng2020deep}. This limitation is particularly noticeable for VRPs characterized by dynamic patterns, e.g., Covering Salesman Problem \cite{Li_Deep_2022}. To solve this problem, Xin et al. \cite{Xin_step_2021} proposed a dynamic embedding method based on AM, which explicitly eliminates the visited nodes and extracts the updated feature embeddings in each decoding step. Similarly, Li et al. \cite{li_learning_2023} proposed a Transformer-based NN to adaptively update the node embeddings with the guidance of the historically constructed solutions. This type of encoder design usually strive for better performance by leveraging more time and computing resources.

In summary, the superior performance of AM has established it as the predominant encoder type. While certain studies considered the demands of specific VRP instances, thus tailoring the Transformer-based encoder design accordingly.

\subsection{Design of Decoding Processes}
\label{sec3.2}

The decoder of most L2C solvers adopts the Autoregressive (AR) approach, selecting a node on the Hamiltonian loop at each decoding step based on the embeddings produced by the encoder. Moreover, the encoders and decoders of these L2C solvers typically have the same network architecture. However, a notable difference lies in the number of network layers between the encoder and decoder. For example, Kool et al. \cite{kool_attention_2019} adopted the attention layer used in the encoder as the backbone of the decoder, with three attention layers in the encoder but only one in the decoder. This design, commonly known as the head heavy encoder and light encoder structure, has been empirically led to high-level performance on various small-scale problems \cite{kwon_pomo_2020, Xin_step_2021, kwon_matrix_2021}. However, it is worth noting that Luo et al. \cite{luo_neural_2023} recently proposed a novel Light Encoder (a single layer of attention) and Heavy Decoder (six attention layers) model (LEHD) to solve large-scale VRPs. LEHD demonstrates reduced sensitivity to the scale of instances and exhibits remarkably improved generalization performance when being applied to large-scale VRP instances.

The sequential nature of the AR approach imposes a fundamental speed limit on these L2C solvers \cite{nowak2018divide}. As a result, these solvers struggle to construct high-quality solutions within short time constraints in real-world scenarios \cite{li_research_2021}. To expedite the decoding process, as shown in Fig~\ref{fig4}, certain L2C solvers \cite{xiao_distilling_2024, xiao2023reinforcement} employ the Non-Autoregressive (NAR) approach, enabling the construction of complete solutions in a single decoding step. For example, Xiao et al. \cite{xiao2023reinforcement} employed the Greedy Search algorithm to decode the outputs of the encoder, i.e., the edge score matrix and the start node pointer, adhering to the order constraints inherent in TSP. Although NAR solvers excel in parallel solution constructions with low decoding latency, they often exhibit inferior performance in terms of quality. Subsequently, to obtain NAR solvers with low decoding latency and the capability to generate high-quality solutions, Xiao et al. \cite{xiao_distilling_2024} incorporated a knowledge distillation method to train a NAR solver (student model) using an AR solver (teacher model). For solvers utilizing NAR decoding methods, advocating for more sophisticated post-processing strategies (see Section~\ref{sec3.4} for more details), rather than solely relying on rudimentary Greedy Search algorithm, demonstrates a promising way of enhancing their performance.

In summary, the sequential nature of AR decoding aligns well with the sequential nature of VRPs, where all nodes are visited sequentially, making AR-type decoders predominant. At the same time, NAR type is still widely adopted, to expedite the decoding process.

\subsection{Markov Decision Process (MDP) of L2C Solvers}
\label{sec3.3}

In this subsection, considering the prevalent adoption of RL among the L2C solvers, we introduce the MDP of L2C solvers. This introduction provides a fresh perspective on understanding the solution construction process of L2C solvers. In addition, the introduction of MDPs of varying solver categories (see Section~\ref{sec4.2} for more details about the MDP of L2I solvers) validates the reasonableness and exclusivity of the taxonomy proposed in this paper. Specifically, we introduce the MDP of L2C solvers from the following five aspects:
\subsubsection{\textbf{State}}
The state $s_t$ at time $t$ encapsulates three kinds of information, namely the local-level, global-level, and current partial solution information \cite{zhang_review_2023}. Specifically, the local-level information is composed of all node-related details, e.g., coordinate and demand. The global-level information encompasses vehicle details, e.g., the remaining capacity of the vehicle for CVRP. The current partial solution information is composed of the previously selected nodes $(a_1, ..., a_{t-1})$, where $a_{t-1}$ denotes the node (action) selected at time $t-1$. In addition, when $t=0$, signifying the initial state, the partial solution is devoid of any selected node.
\subsubsection{\textbf{Action}}
The chosen action $a_t$ at time $t$ denotes a valid node selected from the node set $\bm{V}$. It is imperative to clarify that nodes adhering to the problem constraints, thus not causing violations, are termed valid nodes. For example, in TSP, the previously visited nodes are considered invalid, and the unvisited nodes are regarded as valid. To ensure the selection of valid nodes, those within set $\bm{V}$ that could lead to infeasible solutions or violate constraints undergo masking.

\subsubsection{\textbf{Transition}}
The state transition rule is deterministic, involving the addition of the chosen action $a_t$ to the partial solution, thereby transitioning from $s_{t-1}$ to $s_{t}$. Subsequently, the state undergoes updates to its local-level, global-level, and partial solution information. 

\subsubsection{\textbf{Reward}}
To align the maximization of cumulative rewards with the minimization of the objective value, the reward is typically defined as the negative increment in the objective value (resulting from adding $a_t$ to the current partial solution). Therefore, the value of the cumulative rewards is represented by the negative of the total objective value.
\subsubsection{\textbf{Policy}}
The stochastic policy $\pi$ selects an action according to the probability of actions acquired through NN learning, given the current state $s_t$. This process continues until all nodes are selected, concluding the episode and resulting in the construction of a valid tour.

\subsection{Data Augmentation and Post-processing Strategies}
\label{sec3.4}

In the field of Computer Vision and Natural Language Processing, data augmentation methods have been shown as instrumental in significantly elevating model performance. Similarly, data augmentation methods also play a pivotal role in the field of NCO, which can enrich the training datasets and enhance robustness \cite{kwon_pomo_2020, yao2023dataefficient, luo_neural_2023, jin_pointerformer_2023}. Specifically, to enrich the training datasets, Kwon et al. \cite{kwon_pomo_2020} were the first to introduce the symmetry coordinate transformation into L2C solvers as a data augmentation method. Subsequently, Kim et al. \cite{kim_sym-nco_2022} proposed the orthogonal rotation coordinate transformation method to enrich the training datasets. Following \cite{kwon_pomo_2020, kim_sym-nco_2022}, Yao et al. \cite{yao2023dataefficient} further extended the symmetry of VRPs, and proposed four coordinate transformation methods, namely rotation, symmetry, shrink, and noise. These data augmentation methods enrich the training datasets and remarkably contribute to the high-level performance of L2C solvers. This is particularly evident in the case of SL-based solvers, where a considerable quantity of training instances with high-quality labels is required.

\begin{table*}[!t]
\centering

\caption{\label{table2} Performance of various L2C solves on small-scale TSP and CVRP instances}
	\scalebox{0.7}{
\begin{tabular}{|c|ccc|ccc|ccc|ccc|}
\hline
\multirow{2}{*}{Methods} & \multicolumn{3}{c|}{TSP ($N=50$)}                                                   & \multicolumn{3}{c|}{TSP ($N=100$)}                                                  & \multicolumn{3}{c|}{CVRP ($N=50$)}                                                  & \multicolumn{3}{c|}{CVRP ($N=100$)}                                                \\ \cline{2-13} 
                        & \multicolumn{1}{c|}{Obj.} & \multicolumn{1}{c|}{Gap} & Time                  & \multicolumn{1}{c|}{Obj.} & \multicolumn{1}{c|}{Gap} & Time                  & \multicolumn{1}{c|}{Obj.} & \multicolumn{1}{c|}{Gap} & Time                  & \multicolumn{1}{c|}{Obj.} & \multicolumn{1}{c|}{Gap} & Time                  \\ \hline
Concorde (2007) \cite{applegate_the_2007}$\dagger$			              	       & \multicolumn{1}{c|}{5.696} & \multicolumn{1}{c|}{-} & \multicolumn{1}{c|}{9m} & \multicolumn{1}{c|}{7.765} & \multicolumn{1}{c|}{-} & \multicolumn{1}{c|}{43m} & \multicolumn{1}{c|}{-}     & \multicolumn{1}{c|}{-}    & \multicolumn{1}{c|}{-} & \multicolumn{1}{c|}{-}     & \multicolumn{1}{c|}{-}    & \multicolumn{1}{c|}{-} \\ \hline
HGS (2022) \cite{vidal_hybird_2022}$\dagger$			              	       & \multicolumn{1}{c|}{-} & \multicolumn{1}{c|}{-} & \multicolumn{1}{c|}{-} & \multicolumn{1}{c|}{-} & \multicolumn{1}{c|}{-} & \multicolumn{1}{c|}{-} & \multicolumn{1}{c|}{10.366}     & \multicolumn{1}{c|}{-}    & \multicolumn{1}{c|}{1.2d} & \multicolumn{1}{c|}{15.563}     & \multicolumn{1}{c|}{-}    & \multicolumn{1}{c|}{1.5d} \\ \hline
Ptr-Net-SL (2015) \cite{vinyals_pointer_2015} (greedy)			              	       & \multicolumn{1}{c|}{7.660} & \multicolumn{1}{c|}{34.48\%}    & \multicolumn{1}{c|}{-} & \multicolumn{1}{c|}{}     & \multicolumn{1}{c|}{}    & \multicolumn{1}{c|}{} & \multicolumn{1}{c|}{-}    & \multicolumn{1}{c|}{-}    & \multicolumn{1}{c|}{-} & \multicolumn{1}{c|}{-}    & \multicolumn{1}{c|}{-}    & \multicolumn{1}{c|}{-} \\ \hline
Ptr-Net-RL (2017) \cite{bello_neural_2017}  ($S=1280$)				     	       & \multicolumn{1}{c|}{5.750}     & \multicolumn{1}{c|}{0.95\%} & \multicolumn{1}{c|}{-} & \multicolumn{1}{c|}{8.000}     & \multicolumn{1}{c|}{3.03\%} & \multicolumn{1}{c|}{-} & \multicolumn{1}{c|}{-}     & \multicolumn{1}{c|}{-}    & \multicolumn{1}{c|}{-} & \multicolumn{1}{c|}{-}     & \multicolumn{1}{c|}{-}    & \multicolumn{1}{c|}{-} \\ \hline
S2V-DQN   (2018) \cite{khalil_learning_2017}  (greedy)						     & \multicolumn{1}{c|}{-}     & \multicolumn{1}{c|}{-}    & \multicolumn{1}{c|}{-} & \multicolumn{1}{c|}{8.310}     & \multicolumn{1}{c|}{7.03\%}  & \multicolumn{1}{c|}{-} & \multicolumn{1}{c|}{-}     & \multicolumn{1}{c|}{-}    & \multicolumn{1}{c|}{-} & \multicolumn{1}{c|}{-}     & \multicolumn{1}{c|}{-}    & \multicolumn{1}{c|}{-} \\ \hline
AM (2019) \cite{kool_attention_2019} ($S=1280$)    					                & \multicolumn{1}{c|}{5.730}     & \multicolumn{1}{c|}{0.52\%} & \multicolumn{1}{c|}{-} & \multicolumn{1}{c|}{7.940}     & \multicolumn{1}{c|}{2.26\%} & \multicolumn{1}{c|}{-} & \multicolumn{1}{c|}{10.620}     & \multicolumn{1}{c|}{-}  & \multicolumn{1}{c|}{-} & \multicolumn{1}{c|}{16.230}     & \multicolumn{1}{c|}{-}  & \multicolumn{1}{c|}{-} \\ \hline
POMO (2020)  \cite{kwon_pomo_2020} ($A=8, S=200$)$\dagger$   			           & \multicolumn{1}{c|}{5.696} & \multicolumn{1}{c|}{0.00\%} & \multicolumn{1}{c|}{1.1h} & \multicolumn{1}{c|}{7.770} & \multicolumn{1}{c|}{0.07\%} & \multicolumn{1}{c|}{5.6h} & \multicolumn{1}{c|}{10.397} & \multicolumn{1}{c|}{0.30\%} & \multicolumn{1}{c|}{1.4h} & \multicolumn{1}{c|}{15.672} & \multicolumn{1}{c|}{0.70\%} & \multicolumn{1}{c|}{7.2h} \\ \hline
HAM (2021)  \cite{Zhao_hybird_2021}	($B=10$)									& \multicolumn{1}{c|}{-}     & \multicolumn{1}{c|}{-}    & \multicolumn{1}{c|}{-} & \multicolumn{1}{c|}{-}     & \multicolumn{1}{c|}{-}    & \multicolumn{1}{c|}{-} & \multicolumn{1}{c|}{10.490}     & \multicolumn{1}{c|}{-}    & \multicolumn{1}{c|}{-} & \multicolumn{1}{c|}{17.110}     & \multicolumn{1}{c|}{-}    & \multicolumn{1}{c|}{-} \\ \hline
MDAM (2021) \cite{xin_multi-decoder_2021} { ($B=250$)}          				         & \multicolumn{1}{c|}{5.700} & \multicolumn{1}{c|}{0.03\%} & \multicolumn{1}{c|}{-} & \multicolumn{1}{c|}{7.790} & \multicolumn{1}{c|}{0.38\%} & \multicolumn{1}{c|}{-} & \multicolumn{1}{c|}{10.480} & \multicolumn{1}{c|}{-} & \multicolumn{1}{c|}{-} & \multicolumn{1}{c|}{15.990}& \multicolumn{1}{c|}{-} & \multicolumn{1}{c|}{-} \\ \hline
AM-ASW (2021) \cite{Xin_step_2021} {(greedy)}									 & \multicolumn{1}{c|}{5.760}   & \multicolumn{1}{c|}{1.16\%}  & \multicolumn{1}{c|}{-} & \multicolumn{1}{c|}{8.010}   & \multicolumn{1}{c|}{3.20\%}  & \multicolumn{1}{c|}{-} & \multicolumn{1}{c|}{10.900}     & \multicolumn{1}{c|}{-} & \multicolumn{1}{c|}{-} & \multicolumn{1}{c|}{16.420}     & \multicolumn{1}{c|}{-} & \multicolumn{1}{c|}{-} \\ \hline
POMO-EAS (2022) \cite{hottung_efficient_2022} ($A=8, S=200$)$\dagger$     		       & \multicolumn{1}{c|}{5.696} & \multicolumn{1}{c|}{0.00\%} & \multicolumn{1}{c|}{2h} & \multicolumn{1}{c|}{7.769} & \multicolumn{1}{c|}{0.05\%} & \multicolumn{1}{c|}{10.9h} & \multicolumn{1}{c|}{10.379} & \multicolumn{1}{c|}{0.13\%} & \multicolumn{1}{c|}{3.1h} & \multicolumn{1}{c|}{15.610} & \multicolumn{1}{c|}{0.30\%} & \multicolumn{1}{c|}{16h} \\ \hline
POMO-Sym (2022) \cite{kim_sym-nco_2022} ($A$=8, $S$=200)$\dagger$                 &                                                                                  \multicolumn{1}{c|}{-}     & \multicolumn{1}{c|}{-}    & \multicolumn{1}{c|}{-} & \multicolumn{1}{c|}{7.771} & \multicolumn{1}{c|}{0.08\%} & \multicolumn{1}{c|}{5.6h} & \multicolumn{1}{c|}{-}     & \multicolumn{1}{c|}{-}    & \multicolumn{1}{c|}{-} & \multicolumn{1}{c|}{15.702} & \multicolumn{1}{c|}{0.89\%} & \multicolumn{1}{c|}{7.2h}\\ \hline
POMO-SGBS (2022) \cite{choo_simulation_2022} ($A=8, S=200$) $\dagger$                      & \multicolumn{1}{c|}{-}     & \multicolumn{1}{c|}{-}    & \multicolumn{1}{c|}{-} & \multicolumn{1}{c|}{7.767} & \multicolumn{1}{c|}{0.03\%} & \multicolumn{1}{c|}{1.1d} & \multicolumn{1}{c|}{-}     & \multicolumn{1}{c|}{-}    & \multicolumn{1}{c|}{-} & \multicolumn{1}{c|}{15.579} & \multicolumn{1}{c|}{0.10\%} & \multicolumn{1}{c|}{4.1d}  \\ \hline
Pointerformer (2023) \cite{jin_pointerformer_2023} ($A=8, S=200$)$\dagger$              & \multicolumn{1}{c|}{5.697} & \multicolumn{1}{c|}{0.02\%} & \multicolumn{1}{c|}{1.1h} & \multicolumn{1}{c|}{7.773} & \multicolumn{1}{c|}{0.11\%} & \multicolumn{1}{c|}{5.6h} & \multicolumn{1}{c|}{-}     & \multicolumn{1}{c|}{-}    & \multicolumn{1}{c|}{-} & \multicolumn{1}{c|}{-}     & \multicolumn{1}{c|}{-}    & \multicolumn{1}{c|}{-} \\ \hline
SL-DABL (2023) \cite{yao2023dataefficient} ($A=8, S=200$)                 & \multicolumn{1}{c|}{5.693} & \multicolumn{1}{c|}{0.00\%} & \multicolumn{1}{c|}{-} & \multicolumn{1}{c|}{7.769} & \multicolumn{1}{c|}{0.05\%} & \multicolumn{1}{c|}{-} & \multicolumn{1}{c|}{-}     & \multicolumn{1}{c|}{-}    & \multicolumn{1}{c|}{-} & \multicolumn{1}{c|}{-}     & \multicolumn{1}{c|}{-}    & \multicolumn{1}{c|}{-} \\ \hline
FER-POMO (2023) \cite{li_learning_2023} ($A=8, S=200$)                  & \multicolumn{1}{c|}{-} & \multicolumn{1}{c|}{0.00\%} & \multicolumn{1}{c|}{-} & \multicolumn{1}{c|}{-} & \multicolumn{1}{c|}{0.03\%} & \multicolumn{1}{c|}{-} & \multicolumn{1}{c|}{-}     & \multicolumn{1}{c|}{0.16\%}    & \multicolumn{1}{c|}{-} & \multicolumn{1}{c|}{-}     & \multicolumn{1}{c|}{0.56\%}    & \multicolumn{1}{c|}{-} \\ \hline
LEHD (2023) \cite{luo_neural_2023} ($RRC=100$)                  & \multicolumn{1}{c|}{-} & \multicolumn{1}{c|}{-} & \multicolumn{1}{c|}{-} & \multicolumn{1}{c|}{-} & \multicolumn{1}{c|}{0.01\%} & \multicolumn{1}{c|}{-} & \multicolumn{1}{c|}{-}     & \multicolumn{1}{c|}{-}    & \multicolumn{1}{c|}{-} & \multicolumn{1}{c|}{-}     & \multicolumn{1}{c|}{0.27\%}    & \multicolumn{1}{c|}{-} \\ \hline
\end{tabular}}
\begin{tablenotes}
\item[] Note: Symbol $\dagger$ denotes the results are taken from \cite{ma_learning_2023}, while the rest are taken from the given references. Abbreviations $A$, $S$, and $B$ denote the number of augmentations, samplings, and width of the Beam Search, respectively. In addition, greedy signifies the adoption of Greedy Search during the decoding process. For all experiments on TSP and CVRP, ``Obj.", ``Gap", and ``Time" denote the average objective values, average gaps, and the total run time on the corresponding test dataset with 10k instances, respectively. The gaps are computed relative to the exact algorithm Concorde \cite{applegate_the_2007} for TSP and the SOTA OR algorithm HGS \cite{vidal_hybird_2022} for CVRP. 
\end{tablenotes}
\end{table*}

While L2C solvers can construct solutions within seconds using Greedy Search or Beam Search, they are highly prone to falling into the local optima \cite{xin_multi-decoder_2021}. Specifically, Greedy Search selects the node with the largest probability value at each decoding step, and Beam Search sequentially explores the top $k$ nodes based on their probabilities at each decoding step \cite{vinyals_pointer_2015}. Therefore, other than data augmentation methods, post-processing strategies are also employed to pursue high-quality solutions with a sacrifice of computation time \cite{bello_neural_2017, hottung_efficient_2022, Son_meta_2023,verdu2024scaling}. Initially, in the field of NCO, Bello et al. \cite{bello_neural_2017} proposed two kinds of post-processing strategies, namely Sampling and Active Search (AS). Specifically, Sampling randomly selects nodes utilizing the learned probability during the decoding process, constructs multiple candidate solutions, and ultimately reports the shortest among them. Adopting AS, the model actively updates its parameters for each test instance. Nevertheless, this process of adjusting the weight for each instance is both time-consuming and memory-intensive. To address this problem, Hottung et al. \cite{hottung_efficient_2022} further proposed Effective Active Search (EAS), which selectively adjusts a subset of model parameters. The adoption of EAS remarkably improves the performance of many L2C solvers (e.g., SCA \cite{kim_scale_2022}) without substantially increasing time and memory overhead. Subsequently, to couple with EAS while striving for SOTA performance, Choo et al. \cite{choo_simulation_2022} proposed Simulation Guided Beam Search (SGBS). Specifically, in SGBS, rollouts are performed for the expanded child nodes of Beam Search to identify promising child nodes. Then, the child nodes with poor performance are pruned from consideration, and the beam is expanded to another depth. Recently, Luo et al. \cite{luo_neural_2023} proposed the Random Re-Construct (RRC) post-processing strategy, which resembles the iterative ruin-and-repair approach used in heuristic algorithms. Specifically, RRC randomly samples a partial solution from the constructed complete solution and restructures it using the proposed NCO solver to obtain a new partial solution.

Finally, it is worth mentioning that certain data augmentation (e.g., symmetry coordinate transformation \cite{kwon_pomo_2020}) and post-processing strategies (e.g., EAS \cite{hottung_efficient_2022}) also can potentially be integrated into L2I, L2P-O, and L2P-M solvers to pursue better performance \cite{ma_learning_2023}. For example, Ma et al. \cite{ma_efficient_2022} extended the symmetry coordinate transformation technique, integrating it into L2I solvers to improve the diversity of training sets. Owing to the general applicability of data augmentation and post-processing strategies across all NCO solvers, we refrain from providing repetitive introductions in Sections~\ref{sec4} and \ref{sec5}.

\subsection{Performance of L2C Solvers}
\label{sec3.5}

Table~\ref{table2} presents the performance of different L2C solvers on small-scale TSP and CVRP instances. Before 2019, most studies focused on enhancing the performance of L2C solvers from the perspective of model architecture. Then, with the establishment of AM \cite{kool_attention_2019} and POMO \cite{kwon_pomo_2020} in 2020, the success of these two studies has redirected subsequent research. Instead of exclusively focusing on improving encoders and decoders, there is a distinct shift towards advocating for data augmentation and post-processing strategies (e.g., SGBS \cite{choo_simulation_2022}) to enhance the effectiveness of AM and POMO. Through the incorporation of advanced data augmentation and post-processing strategies, current L2C solvers still showcase performance on par with OR algorithms. For example, SGBS \cite{choo_simulation_2022} attains gap values of 0.03\% and 0.1\% on the 100-node TSP and 100-node CVRP, respectively. However, existing L2C solvers have certain inadequacies when with faced specific situations. First, it is worth mentioning that certain L2C solvers prioritized high performance over solving time, leading to the conventional OR algorithms achieving shorter solution time \cite{xiao2023reinforcement}. This contradicts the inherent advantage of NCO solvers, i.e., their ability to solve VRP instances rapidly.  Additionally, the generalization capability of these L2C solvers needs to be enhanced to better solve instances with different data distribution patterns and problem scales (see Section~\ref{sec6.1} for more details). Finally, due to the time and memory complexity, existing Transformer-based L2C solvers struggle with handling large-scale VRP instances, e.g., the 10,000-node TSP (see Section~\ref{sec6.2} for more details).

\section{Learning to Improve (L2I) Solvers} 
\label{sec4}

\begin{figure}[!t]
\centering
	\includegraphics[scale=0.8]{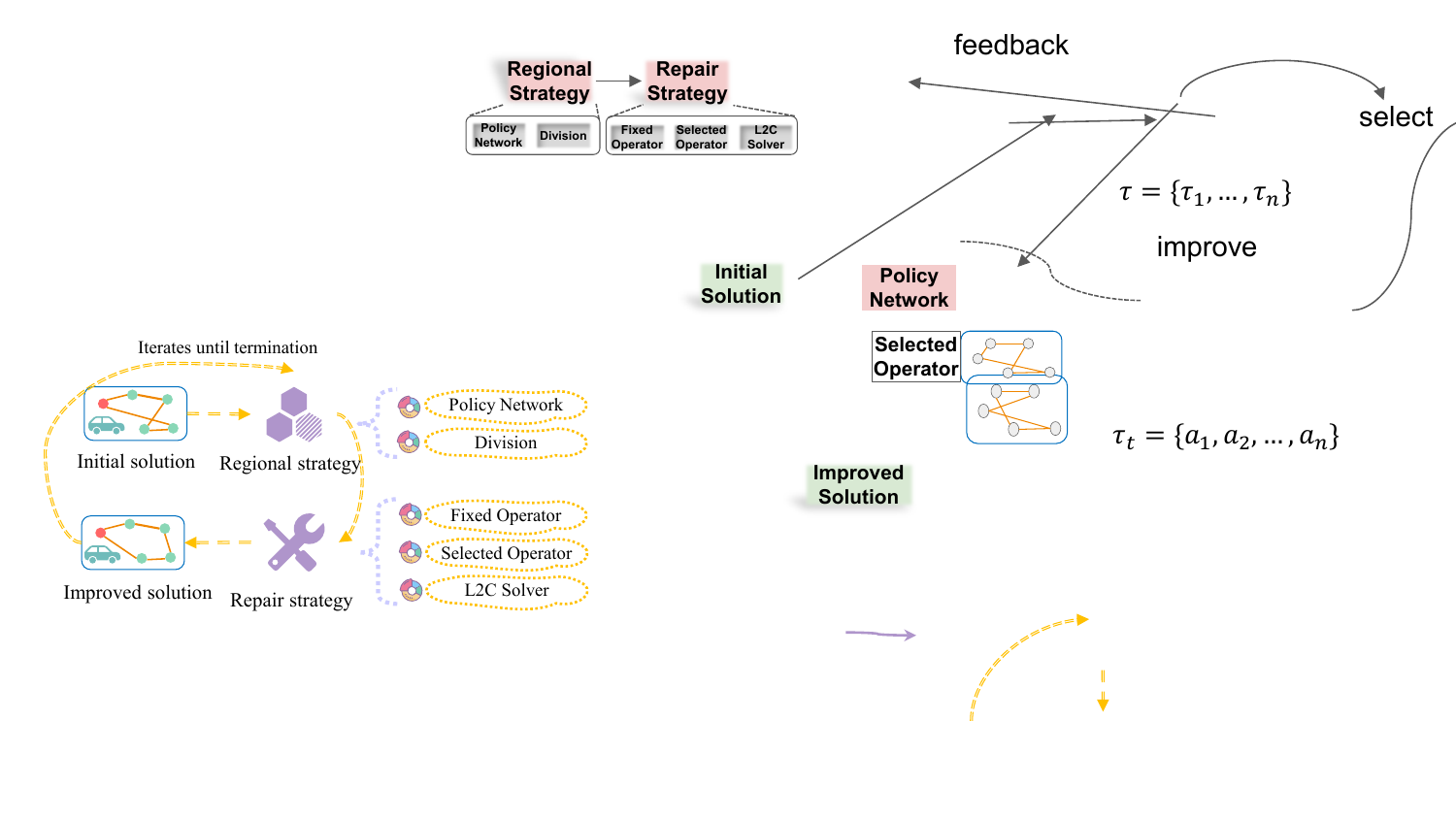}
	\caption{Illustration of the iterative improvement solutions process of L2I solvers, starting from an initial complete solution and ending within a given timeframe. L2I solvers first rely on regional strategies to select regions (typically node pairs). Subsequently, diverse strategies are employed to repair the sub-tours of the selected regions.} 
	\label{fig5}
\end{figure}

Inspired by the idea of ruining and repairing in heuristic algorithms, L2I solvers iteratively improve current complete solutions, facilitating the exploration of \mbox{(sub-)}optimal solutions within a specified time frame, as shown in Fig.~\ref{fig5}. Specifically, at each time step, the regional strategy of L2I solvers identifies the regions (typically node pairs) requiring ``repair". Subsequently, L2I solvers exploit the repair strategy in the selected regions to improve current complete solutions. To thoroughly review L2I solvers, Sections~\ref{sec4.1} and \ref{sec4.2} outline the designs of regional strategy and three repair strategies, respectively. Subsequently, Sections~\ref{sec4.3} and \ref{sec4.4} introduce the MDP of L2I and provide a comprehensive overview of the recent L2I solvers' performance on small-scale TSP and CVRP instances, respectively.

\subsection{Design of Regional Strategies}
\label{sec4.1}

L2I solvers generally use a policy network to identify the regions that require repair (in another word, performance elevation) based on current solutions \cite{Zhou_Learning_2023}. In the field of NCO, Chen and Tian \cite{chen_learning_2019} first adopted a fully connected NN as a score predictor to identify a few nodes where rewriting might be beneficial. Subsequently, Costa et al. \cite{da_costa_learning_2020} and Sui et al. \cite{sui_learning_2021} adopted a network block consisting of GNN and RNN to select nodes based on current solutions. Considering the superior performance of Transformer-based networks for VRP, Wu et al. \cite{wu_learning_2022} proposed a Transformer-based network to choose node pairs for executing the 2-opt operator (see Section~\ref{sec4.2} for more details about 2-opt operator). Subsequently, Ma et al. \cite{ma_learning_2021} proposed DACT to employ two Transformer networks to individually learn information from node coordinates and current solutions. Such information is then collaboratively used to determine the selection of node pairs. However, due to the adoption of two Transformer-based networks, the computing efficiency of DACT is not satisfactory \cite{zhang_review_2023}. To improve computing efficiency, Ma et al. \cite{ma_efficient_2022} further proposed the Synthesis Attention mechanism, capable of simultaneously learning embeddings of node coordinates and current solutions to attain a comprehensive representation. While this regional strategy typically yields satisfactory results for small-scale VRPs, its efficiency diminishes when used to solve large-scale VRPs. This is because using NNs to select regions consisting of only a few nodes in large-scale VRPs could exponentially increase the overall solution time, ultimately leading to inefficient solution provision.	

To better tackle large-scale VRP instances, certain L2I solvers \cite{ ye_glop_2024} propose a division strategy inspired by the D\&C-based manner, rather than relying on NNs to select regions. Specifically, these D\&C-based L2I solvers utilize the visiting order of the current complete solution to divide the problem into multiple sub-problems. Subsequently, all these sub-problems are simultaneously entered into the repair strategy for performance elevation. Finally, only the sub-problems showing improvement are considered for iteratively updating the current solutions. Moreover, to simultaneously improve multiple sub-problems, these L2I solvers \cite{cheng_select_2023, ye_glop_2024} also design a special repair strategy (see Section~\ref{sec4.2} for more details). For these solvers, to prevent increased distance costs during the merging of sub-problems, establishing an efficient division strategy is crucial.

In summary, numerous L2I solvers favor Transformer-based NNs to identify regions necessitating repair in small-scale VRP instances. In addition, to handle large-scale VRP instances, certain L2I solvers divide the problem into multiple sub-problems and strive to improve all sub-problems.

\subsection{Design of Repair Strategies}
\label{sec4.2}
\begin{figure}[!t]
\centering
	\includegraphics[scale=0.35]{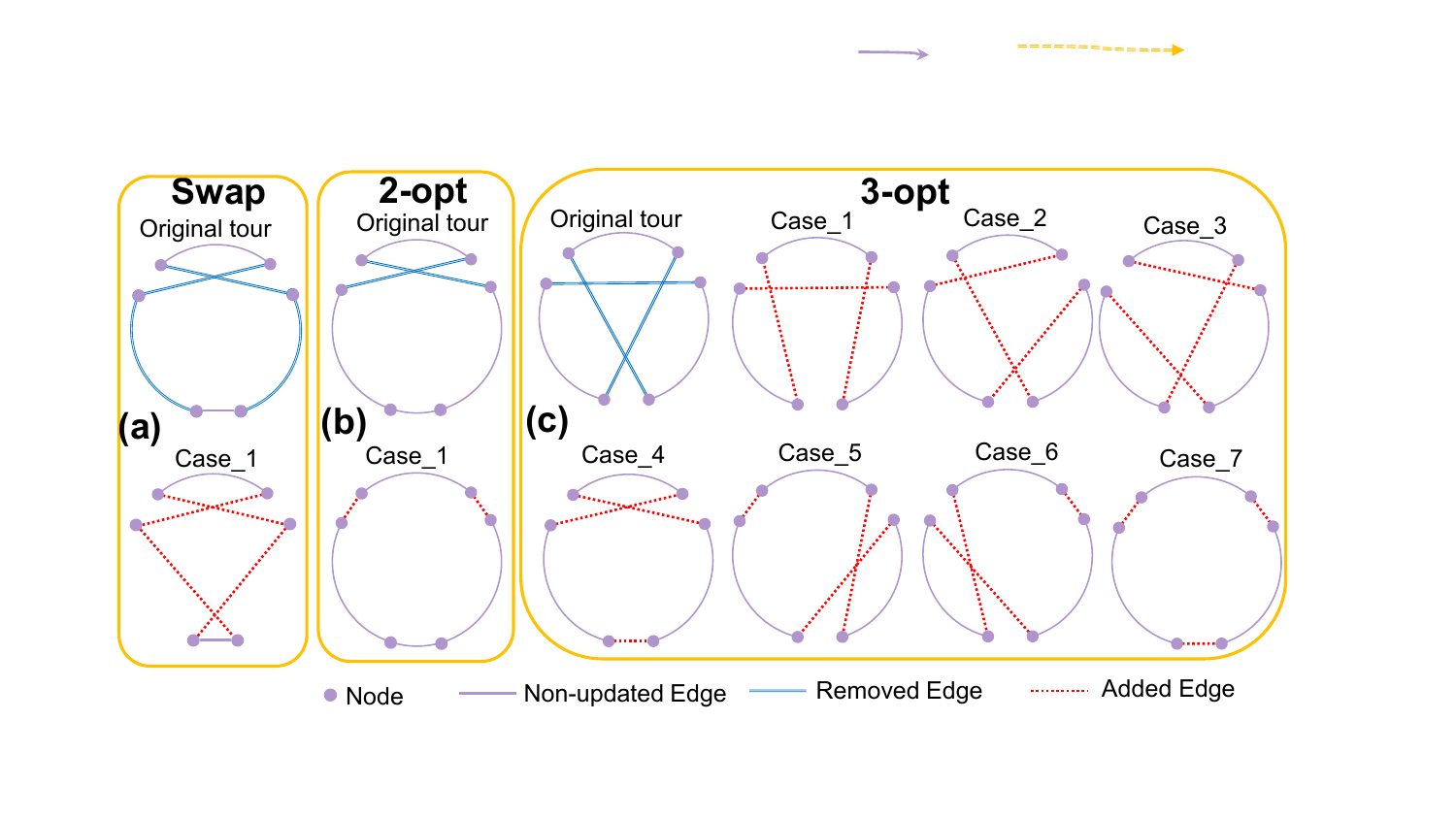}
	\caption{Illustration of the exchange processes of swap, 2-opt, and 3-opt operators. The blue and red links denote the edges selected to be removed and selected to be added from the current tour, respectively. For swap and 2-opt operators, only one exchange takes place. However, for a 3-opt operator, seven different exchanges take place.} 
	\label{fig6}
\end{figure}

Currently, L2I solvers employ one of three distinct repair strategies to repair sub-tours of the selected regions. Initially, for simplicity, certain L2I solvers exploit the fixed operator throughout the entire iteration process to improve current solutions. For example, Chen and Tian \cite{chen_learning_2019} adopted swap operator to iteratively improve current solutions, involving the exchange of the visiting order of two nodes, as shown in Fig.~\ref{fig6}(a). Subsequently, Wu et al. \cite{wu_learning_2022} and Costa et al. \cite{da_costa_learning_2020} adopted \mbox{2-opt} operator for iterative improvements in current solutions. As shown in Fig.~\ref{fig6}(b), the {2-opt} operator reconnects two new edges at a time. Despite the simplicity of swap and 2-opt operators, their performance falls short in searching for (sub-)optimal solutions within a limited timeframe (see Section~\ref{sec4.4}).

To pursue better performance, certain L2I solvers exploit additional NNs to choose suitable operators for the regions selected by the regional strategy. For example, Lu et al. \cite{lu_learning-based_2020} exploited a Transformer-based network to select operators from two superior sets customized for CVRP, namely intra-route and inter-route sets. Specifically, intra-route operators endeavour to alter the order of visiting nodes within a single route, while inter-route ones attempt to exchange nodes among different routes. Meanwhile, Sui et al. \cite{sui_learning_2021} replaced 2-opt with 3-opt and proposed an NN to choose from seven different cases of the 3-opt operator, as depicted in Fig.~\ref{fig6}(c). Generally speaking, the operator's performance improves with a larger value of $k$ in $k$-opt operators, but simultaneously, the computing complexity significantly increases. To automatically determine the values of $k$ at different iteration stages, Ma et al. \cite{ma_learning_2023} proposed NeuOpt, which decomposes the sequential $k$-opt operator into a structured sequence consisting of a starting move, $k$ intermediate moves (determined by the proposed repair network), and an ending move. This design effectively balances coarse-grained (larger $k$) and fine-grained (smaller $k$) search strategies.

Finally, to tackle large-scale VRP instances, certain L2I solvers (e.g., SO \cite{cheng_select_2023} and GLOP \cite{ye_glop_2024}) exploit L2C solvers (e.g., AM \cite{kool_attention_2019}) to directly repair all sub-problems, bypassing operators like the 2-opt operator. The reason for employing L2C solvers as the repair strategy lies in the regional strategy utilized by these L2I solvers. These solvers divide large-scale problems into multiple smaller sub-problems, which can be efficiently solved by L2C solvers in parallel. In contrast, traditional operators lack the capability to simultaneously solve a batch of sub-problems. In addition, Li et al. \cite{li2025destroy} proposed DRHG, which first decomposes the complete solution into certain isolated nodes and segments (with each segment composed of multiple consecutive nodes). Subsequently, DRHG simplifies the segments into hyperedges (temporally retaining only the two endpoints) and utilizes the L2C solver to connect the endpoints of these hyperedges with the isolated points, ultimately reconstructing a complete solution. This design directs the L2C solver’s attention to the disrupted regions while reducing the encoding complexity of all nodes within the segments, particularly suited to handle large-scale VRP instances. In addition, it is worth mentioning that while these solvers employ L2C solvers as their repair strategy, their key objective remains as iteratively improving the current complete solutions. Hence, we categorize these solvers into L2I categories.

\begin{table*}[!t]
\centering
\caption{\label{table3} Performance of various L2I solves on small-scale TSP and CVRP instances}
	\scalebox{0.7}{
\begin{tabular}{|c|ccc|ccc|ccc|ccc|}
\hline
\multirow{2}{*}{Methods} & \multicolumn{3}{c|}{TSP ($N=50$)}                                                   & \multicolumn{3}{c|}{TSP ($N=100$)}                                                  & \multicolumn{3}{c|}{CVRP ($N=50$)}                                                  & \multicolumn{3}{c|}{CVRP ($N=100$)}                                                \\ \cline{2-13} 
                        & \multicolumn{1}{c|}{Obj.} & \multicolumn{1}{c|}{Gap} & Time                  & \multicolumn{1}{c|}{Obj.} & \multicolumn{1}{c|}{Gap} & Time                  & \multicolumn{1}{c|}{Obj.} & \multicolumn{1}{c|}{Gap} & Time                  & \multicolumn{1}{c|}{Obj.} & \multicolumn{1}{c|}{Gap} & Time                  \\ \hline
Concorde (2007) \cite{applegate_the_2007}$\dagger$			              	       & \multicolumn{1}{c|}{5.696} & \multicolumn{1}{c|}{-} & \multicolumn{1}{c|}{9m} & \multicolumn{1}{c|}{7.765} & \multicolumn{1}{c|}{-} & \multicolumn{1}{c|}{43m} & \multicolumn{1}{c|}{-}     & \multicolumn{1}{c|}{-}    & \multicolumn{1}{c|}{-} & \multicolumn{1}{c|}{-}     & \multicolumn{1}{c|}{-}    & \multicolumn{1}{c|}{-} \\ \hline
HGS (2022) \cite{vidal_hybird_2022}$\dagger$			              	       & \multicolumn{1}{c|}{-} & \multicolumn{1}{c|}{-} & \multicolumn{1}{c|}{-} & \multicolumn{1}{c|}{-} & \multicolumn{1}{c|}{-} & \multicolumn{1}{c|}{-} & \multicolumn{1}{c|}{10.366}     & \multicolumn{1}{c|}{-}    & \multicolumn{1}{c|}{1.2d} & \multicolumn{1}{c|}{15.563}     & \multicolumn{1}{c|}{-}    & \multicolumn{1}{c|}{1.5d} \\ \hline
NeuRewriter (2019) \cite{chen_learning_2019} ($T=40K$)	              	       & \multicolumn{1}{c|}{-}     & \multicolumn{1}{c|}{-}    & \multicolumn{1}{c|}{-} & \multicolumn{1}{c|}{-}     & \multicolumn{1}{c|}{-}    & \multicolumn{1}{c|}{-} & \multicolumn{1}{c|}{10.510}     & \multicolumn{1}{c|}{-}    & \multicolumn{1}{c|}{-} & \multicolumn{1}{c|}{16.100}     & \multicolumn{1}{c|}{-}    & \multicolumn{1}{c|}{-} \\ \hline
N2I (2020) \cite{lu_learning-based_2020} ($T=40K$)                                      & \multicolumn{1}{c|}{-}     & \multicolumn{1}{c|}{-}    & \multicolumn{1}{c|}{-} & \multicolumn{1}{c|}{-}     & \multicolumn{1}{c|}{-}    & \multicolumn{1}{c|}{-} & \multicolumn{1}{c|}{10.350}     & \multicolumn{1}{c|}{-}    & \multicolumn{1}{c|}{-} & \multicolumn{1}{c|}{15.570}     & \multicolumn{1}{c|}{-}    & \multicolumn{1}{c|}{-} \\ \hline
NLNS (2020) \cite{hottung_neural_2020} ($T=5K$) $\dagger$                   & \multicolumn{1}{c|}{-}     & \multicolumn{1}{c|}{-}    & \multicolumn{1}{c|}{-} & \multicolumn{1}{c|}{-}     & \multicolumn{1}{c|}{-}    & \multicolumn{1}{c|}{-} & \multicolumn{1}{c|}{10.506} & \multicolumn{1}{c|}{1.35\%} & \multicolumn{1}{c|}{1.4h} & \multicolumn{1}{c|}{15.915} & \multicolumn{1}{c|}{2.26\%} & \multicolumn{1}{c|}{2.4h} \\ \hline
N2OPT (2020) \cite{da_costa_learning_2020} ($T=2K$) $\dagger$ 	         	       & \multicolumn{1}{c|}{5.703}  & \multicolumn{1}{c|}{0.12\%}  & \multicolumn{1}{c|}{40m} & \multicolumn{1}{c|}{7.824}  & \multicolumn{1}{c|}{0.77\%}  & \multicolumn{1}{c|}{1.1h} & \multicolumn{1}{c|}{-}     & \multicolumn{1}{c|}{-}    & \multicolumn{1}{c|}{-} & \multicolumn{1}{c|}{-}     & \multicolumn{1}{c|}{-}    & \multicolumn{1}{c|}{-} \\ \hline
N3OPT (2021) \cite{sui_learning_2021} ($T=2K$) $\dagger$  	              	       & \multicolumn{1}{c|}{5.700}  & \multicolumn{1}{c|}{0.08\%} & \multicolumn{1}{c|}{48m} & \multicolumn{1}{c|}{7.820}  & \multicolumn{1}{c|}{0.74\%} & \multicolumn{1}{c|}{1.3h}& \multicolumn{1}{c|}{-}     & \multicolumn{1}{c|}{-}    & \multicolumn{1}{c|}{-} & \multicolumn{1}{c|}{-}     & \multicolumn{1}{c|}{-}    & \multicolumn{1}{c|}{-} \\ \hline
MT (2021) \cite{wu_learning_2022}  ($T=5K$) $\dagger$ 			              	       & \multicolumn{1}{c|}{5.709} & \multicolumn{1}{c|}{0.23\%}  & \multicolumn{1}{c|}{1.3h} & \multicolumn{1}{c|}{7.884} & \multicolumn{1}{c|}{1.54\%}  & \multicolumn{1}{c|}{2h} & \multicolumn{1}{c|}{10.544} & \multicolumn{1}{c|}{1.72\%} & \multicolumn{1}{c|}{4.2h} & \multicolumn{1}{c|}{16.165} & \multicolumn{1}{c|}{3.87\%} & \multicolumn{1}{c|}{5h} \\ \hline
DACT (2021) \cite{ma_learning_2021} ($T=10K$) $\dagger$ 	              	       & \multicolumn{1}{c|}{5.696} & \multicolumn{1}{c|}{0.00\%} & \multicolumn{1}{c|}{2.7h} & \multicolumn{1}{c|}{7.772} & \multicolumn{1}{c|}{0.10\%} & \multicolumn{1}{c|}{7h} & \multicolumn{1}{c|}{10.383} & \multicolumn{1}{c|}{0.16\%} & \multicolumn{1}{c|}{16h} & \multicolumn{1}{c|}{15.736} & \multicolumn{1}{c|}{1.11\%} & \multicolumn{1}{c|}{1.6d} \\ \hline
NeuOpt (2023) \cite{ma_learning_2023} ($T=10K$) $\dagger$              	       & \multicolumn{1}{c|}{5.696} & \multicolumn{1}{c|}{0.00\%} & \multicolumn{1}{c|}{1.1h} & \multicolumn{1}{c|}{7.766} & \multicolumn{1}{c|}{0.02\%} & \multicolumn{1}{c|}{2.8h} &
\multicolumn{1}{c|}{10.375}  & \multicolumn{1}{c|}{0.08\%} & \multicolumn{1}{c|}{2h} & \multicolumn{1}{c|}{15.656}  & \multicolumn{1}{c|}{0.60\%} & \multicolumn{1}{c|}{4.6h}  \\ \hline
DRHG (2025) \cite{li2025destroy} ($T=1K$)     & \multicolumn{1}{c|}{-} & \multicolumn{1}{c|}{-} & \multicolumn{1}{c|}{-} & \multicolumn{1}{c|}{} & \multicolumn{1}{c|}{0.00\%} & \multicolumn{1}{c|}{-} &
\multicolumn{1}{c|}{-}  & \multicolumn{1}{c|}{-} & \multicolumn{1}{c|}{-} & \multicolumn{1}{c|}{-}  & \multicolumn{1}{c|}{-0.02\%} & \multicolumn{1}{c|}{-}  \\ \hline

\end{tabular}}
\begin{tablenotes}
\item[] Note: Symbol $\dagger$ denotes the results are taken from \cite{ma_learning_2023}, while the rest are taken from the given references. Abbreviation $T$ denotes the number of iterations. For all experiments on TSP and CVRP, ``Obj.", ``Gap", and ``Time" denote the average objective values, average gaps, and the total run time on the corresponding test dataset with 10k instances, respectively. The gaps are computed relative to the exact algorithm Concorde \cite{applegate_the_2007} for TSP and the SOTA OR algorithm HGS \cite{vidal_hybird_2022} for CVRP. 
\end{tablenotes}
\end{table*}
In summary, to pursue better performance, many recent L2I solvers employ NNs to select suitable operators instead of relying on a fixed single one. Because of the requirement for improving multiple sub-problems in parallel, certain L2I solvers incorporate L2C solvers to repair them.

\subsection{Markov Decision Process (MDP) of L2I Solvers}
\label{sec4.3}
In this subsection, to elucidate the fundamental distinction between L2C solvers and L2I solvers, we introduce the MDP of L2I solvers from the following aspects:
\subsubsection{\textbf{State}}
The state $s_t$ reflects three kinds of information, namely local-level, global-level, and current solution information at time $t$. Local-level information encompasses details about all nodes (e.g., coordinates) \cite{zhang_review_2023}. The global-level information consists of historical solutions and their corresponding costs. Unlike the MDP of L2C solvers, the current solution information $\delta_{t-1}$ of the MDP for L2I solvers is a complete solution. In addition, when $t=0$, the initial solution $\delta_0$ can be any randomly generated feasible solution.
\subsubsection{\textbf{Action}}
The action $a_t$ represents a predefined operation that transforms the current solution into an alternative one. For example, one form of action involves specifying nodes to perform a pairwise local search operation (e.g., 2-opt operator) in \cite{ma_learning_2021, wu_learning_2022}. Analogous to L2C solvers, ensuring a consistently feasible solution involves applying mask mechanism to the action space in L2I solvers.

\subsubsection{\textbf{Transition}}
In numerous L2I solvers, the state transition involves updating the current solution $\delta_{t-1}$ with the improved solution $\delta_t$. Simultaneously, the information at both the local and global levels of the state undergoes corresponding updates.
\subsubsection{\textbf{Reward}}
To ensure that cumulative rewards accurately capture the overall improvement, i.e., a lower objective value from the initial solution, the reward is commonly defined as the immediate reduction in the objective value of the current best solution following action $a_t$.
\subsubsection{\textbf{Policy}}
The policy $\pi$ of the MDP for L2I solvers is parameterized by an NN in most cases. Unlike L2C solvers, the number of iterations of the L2I solvers $T$ could be any predefined value. Therefore, if $T$ varies, comparing different L2I solvers becomes inequitable. Finally, the best solution identified across the entire iteration process is reported as the final solution.

\subsection{Performance of L2I Solvers}
\label{sec4.4}

Table~\ref{table3} presents the performance of various L2I solvers on small-scale TSP and CVRP instances. Notably, NeuOpt \cite{ma_learning_2023}, the most novel L2I solver utilizing the $k$-opt operator, outperforms the others, demonstrating only 0.02\% and 0.6\% gaps compared to Concorde and HGS, respectively. However, for L2I solvers, it is crucial to acknowledge that the problem of unfair comparisons becomes more pronounced when employing varying number of iterations $T$. For a fair comparison, adopting a fixed number of iterations for all L2I solvers is imperative \cite{liu_how_2023}. In addition, the solution time also undergoes a significant rise with an increase of $T$. Finally, existing L2I solvers typically begin with randomly initiated feasible solutions and iteratively improve them for straightforwardness. However, this random initialization strategy hinders convergence and increases search overhead, particularly for large-scale VRP instances \cite{Taillard_POPMUSIC_2019}. Future studies should explore more efficient strategies borrowed from the heuristic algorithms for constructing initial solutions to improve the time efficiency of L2I solvers.

\section{Learning to Predict (L2P) Solvers} 
\label{sec5}

\begin{figure}[!t]
\centering
	\includegraphics[scale=0.65]{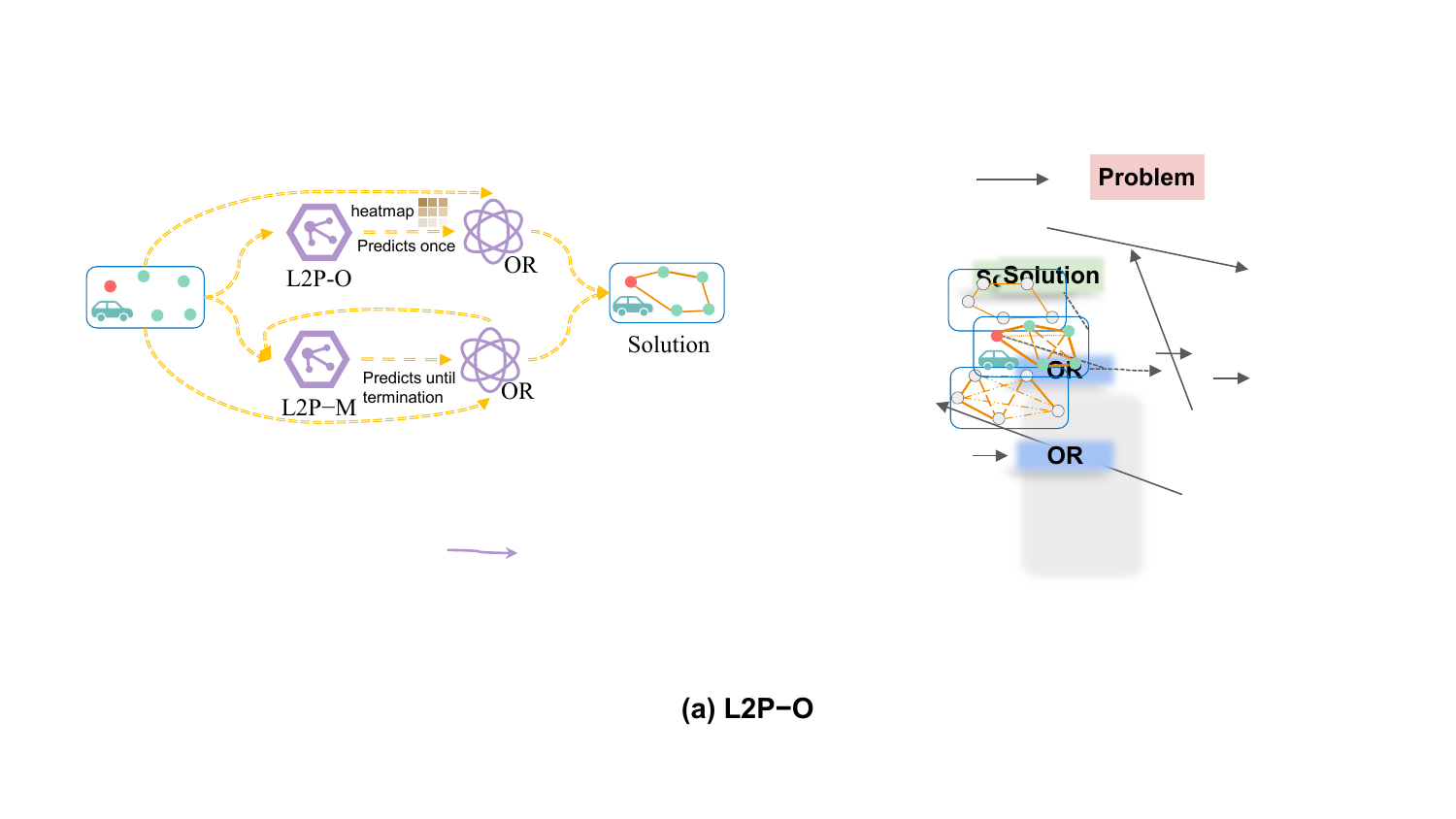}
	\caption{Illustration of the information prediction processes of L2P-O and L2P-M solvers. L2P-O solvers aim to enhance OR algorithms by providing valuable information solely before the searches begin. Subsequently, OR algorithms solve VRP instances by leveraging the predicted information along with problem definitions. Conversely, L2P-M solvers collaborate with OR algorithms continuously during the search processes, offering prediction of key information at each decision step. These L2P-M solvers take states of the OR algorithms, encompassing problem definitions, as their inputs \cite{zheng_combining_2021}.}
	\label{fig7}
\end{figure}

In many cases, relying solely on NNs to address VRPs may not be the optimal strategy. Instead, there is an emerging advocacy for integrating NNs with OR algorithms, as depicted in Fig.~\ref{fig7}. This integration aims to expedite the search process of OR algorithms and enhance the quality of solutions by leveraging DL methods to predict crucial information \cite{bengio_machine_2021, dernedde2024moco, kim2024ant}. Sections~\ref{sec5.1} and \ref{sec5.2} comprehensively review L2P-O and L2P-M solvers, respectively. Subsequently, Section~\ref{sec5.3} presents the performance of existing L2P-O and L2P-M solvers on small-scale VRP and CVRP instances.

\subsection{Learning to Predict-Once (L2P-O) Solvers}
\label{sec5.1}

A group of L2P-O solvers predict information only once before commencing searches with OR algorithms, aiming to expedite the search processes. For example, to confine the search space and thus expedite the search processes of DP, Kool et al. \cite{kool_deep_2022} adopted a Graph Convolution Networks (GCN) to predict a sparse heatmap for each TSP instance, which highlights promising edges. Subsequently, DP excludes solutions with fewer ``heated" edges, effectively eliminating a multitude of inappropriate solutions and confining the search space, thereby reducing runtime. Similar to \cite{kool_deep_2022}, other studies \cite{sun_generalization_2021, prates_learning_2019, Sun_Using_2021} also employed L2P-O solvers to predict the scores of edges, effectively confining the search space and accelerating the search processes. In addition, it is worth mentioning that certain studies \cite{joshi_efficient_2019, hudson_graph_2022, goh2022combining, min2023unsupervised} adopted search algorithms, e.g., Monte Carlo Tree Search (MCTS) and Beam Search, rather than OR algorithms, to construct solutions based on the predicted heatmaps. Nevertheless, because the solution construction process of these methods employing search algorithms (e.g., \cite{joshi_efficient_2019}) is closely similar to OR algorithms (e.g., \cite{kool_deep_2022}), we categorize these methods as L2P-O solvers as well. For example, Joshi et al. \cite{joshi_efficient_2019} leveraged GCN to predict heatmaps and employed Beam Search on the resulting heatmaps to construct solutions. Specifically, they formulated TSP instances as 2D Euclidean graphs and utilized the optimal solutions as labels for training GCN to predict heatmaps. While constructing solutions using search algorithms is more efficient compared to the OR algorithms, it comes with the inadequacy of being unable to solve multi-constrained VRPs.

Meanwhile, other L2P-O solvers predict information to derive higher-quality solutions. For instance, Xin et al. \cite{xin_neurolkh_2021} trained a GNN for predicting edge scores and node penalties, resulting in a more potent edge candidate set than the one created by the original Minimum Spanning Tree used in LKH \cite{helsgaun_effective_2000}. More recently, Ye et al. \cite{ye_deepaco_2023} proposed Deep Ant Colony Optimization (DeepACO), which substitutes human-designed heuristic measures (e.g., Euclidean distances of nodes) in ACO with superior heuristic measures predicted by GNN. Finally, certain methods exploit NNs to configure parameters for OR algorithms or select the optimal OR algorithm for each instance, significantly improving the quality of the derived solutions \cite{karimi_machine_2022, kerschke_leveraging_2018, gutierrez_selecting_2019, qin_novel_2021}. For example, Gutierrez-Rodríguez et al. \cite{gutierrez_selecting_2019} defined a set of meta-features that appropriately characterize different VRP instances. Subsequently, they employed a fully connected NN to predict the most suitable OR algorithm for each instance.

In summary, certain L2P-O solvers exploit the predicted information to expedite the search processes of OR algorithms, while others leverage such information to enhance the quality of solutions generated by OR algorithms. Meanwhile, in addition to the advantages aforementioned, it should not be overlooked that L2P-O solvers also face the generalization capability inadequacy, similar to L2C and L2I solvers.
\subsection{Learning to Predict-Multiplicity (L2P-M) Solvers}
\label{sec5.2}

Compared with L2P-O solvers, L2P-M ones exploit NNs to make the same type of decisions multiple times, aiming to enhance the performance of conventional OR algorithms in terms of time complexity. For example, to expedite search processes, Zheng et al. \cite{zheng_combining_2021} substituted the inflexible process of traversing the candidate set to select edges in the $k$-opt operator of LKH \cite{helsgaun_effective_2000}. Instead, an RL-based method is employed to predict key information and guide the decision-making process for edge selections within the candidate set. Meanwhile, certain \mbox{L2P-M} solvers accelerate the search processes of DP \cite{yang2018boosting, xu_deep_2020}. For example, Xu et al. \cite{xu_deep_2020} introduced a general framework called NDP, which exploits fully connected NNs to replace DP's value or policy functions at each decision step. It is worth mentioning that although having the same purpose as \cite{kool_deep_2022}, the L2P-M solvers \cite{yang2018boosting} and \cite{xu_deep_2020} exploited the NN to predict multiple pieces of information for each TSP instance in the whole process of DP. In contrast, the L2P-O solver DPDP \cite{kool_deep_2022} produces heatmaps only once for each TSP instance. Hence, the proposed taxonomy is still applicable in this case to exclusively divide these NCO solvers into respective types.

\begin{table*}[!t]
\centering

\caption{\label{table4} Performance of various L2P solves on small-scale TSP and CVRP instances}
	\scalebox{0.7}{
\begin{tabular}{|c|c|ccc|ccc|ccc|ccc|}
\hline
\multirow{2}{*}{Methods} &\multirow{2}{*}{Type} & \multicolumn{3}{c|}{TSP ($N=50$)}                                                   & \multicolumn{3}{c|}{TSP ($N=100$)}                                                  & \multicolumn{3}{c|}{CVRP ($N=50$)}                                                  & \multicolumn{3}{c|}{CVRP ($N=100$)}                                                \\ \cline{3-14} & &
\multicolumn{1}{c|}{Obj.} & \multicolumn{1}{c|}{Gap} & Time                  & \multicolumn{1}{c|}{Obj.} & \multicolumn{1}{c|}{Gap} & Time                  & \multicolumn{1}{c|}{Obj.} & \multicolumn{1}{c|}{Gap} & Time                  & \multicolumn{1}{c|}{Obj.} & \multicolumn{1}{c|}{Gap} & Time                  \\ \hline
Concorde (2007) \cite{applegate_the_2007}$\dagger$   &OR		              	       & \multicolumn{1}{c|}{5.696} & \multicolumn{1}{c|}{-} & \multicolumn{1}{c|}{9m} & \multicolumn{1}{c|}{7.765} & \multicolumn{1}{c|}{-} & \multicolumn{1}{c|}{43m} & \multicolumn{1}{c|}{-}     & \multicolumn{1}{c|}{-}    & \multicolumn{1}{c|}{-} & \multicolumn{1}{c|}{-}     & \multicolumn{1}{c|}{-}    & \multicolumn{1}{c|}{-} \\ \hline
HGS (2022) \cite{vidal_hybird_2022}$\dagger$			&OR            	       & \multicolumn{1}{c|}{-} & \multicolumn{1}{c|}{-} & \multicolumn{1}{c|}{-} & \multicolumn{1}{c|}{-} & \multicolumn{1}{c|}{-} & \multicolumn{1}{c|}{-} & \multicolumn{1}{c|}{10.366}     & \multicolumn{1}{c|}{-}    & \multicolumn{1}{c|}{1.2d} & \multicolumn{1}{c|}{15.563}     & \multicolumn{1}{c|}{-}    & \multicolumn{1}{c|}{1.5d} \\ \hline
GCN (2019) \cite{joshi_efficient_2019}$\dagger$	 &L2P-O/BS		          	       & \multicolumn{1}{c|}{5.698} & \multicolumn{1}{c|}{0.04\%} & \multicolumn{1}{c|}{23m} & \multicolumn{1}{c|}{7.869} & \multicolumn{1}{c|}{1.35\%} & \multicolumn{1}{c|}{46m} & \multicolumn{1}{c|}{-}     & \multicolumn{1}{c|}{-}    & \multicolumn{1}{c|}{-} & \multicolumn{1}{c|}{-}     & \multicolumn{1}{c|}{-}    & \multicolumn{1}{c|}{-} \\ \hline
Att-GCN (2021) \cite{fu_generalize_2021}$\dagger$	&L2P-O/MCTS            	       & \multicolumn{1}{c|}{5.691} & \multicolumn{1}{c|}{0.01\%} & \multicolumn{1}{c|}{8m} & \multicolumn{1}{c|}{7.764} & \multicolumn{1}{c|}{0.04\%} & \multicolumn{1}{c|}{15m} & \multicolumn{1}{c|}{-}     & \multicolumn{1}{c|}{-}    & \multicolumn{1}{c|}{-} & \multicolumn{1}{c|}{-}     & \multicolumn{1}{c|}{-}    & \multicolumn{1}{c|}{-} \\ \hline
CAVE-Opt-DE (2021) \cite{hottung_learning_2021}$\dagger$   &L2P-M/DE                 & \multicolumn{1}{c|}{-}     & \multicolumn{1}{c|}{0.02\%}    & \multicolumn{1}{c|}{2.5d} & \multicolumn{1}{c|}{-}     & \multicolumn{1}{c|}{0.34\%}    & \multicolumn{1}{c|}{1.8d} & \multicolumn{1}{c|}{10.400}     & \multicolumn{1}{c|}{-}    & \multicolumn{1}{c|}{4.7d} & \multicolumn{1}{c|}{15.750}     & \multicolumn{1}{c|}{-}    & \multicolumn{1}{c|}{11d} \\ \hline
VSR-LKH (2021) \cite{zheng_combining_2021}    		&L2P-M/LKH          	       & \multicolumn{1}{c|}{-}     & \multicolumn{1}{c|}{-}    & \multicolumn{1}{c|}{-} & \multicolumn{1}{c|}{7.765}     & \multicolumn{1}{c|}{0.01\%}  & \multicolumn{1}{c|}{-} & \multicolumn{1}{c|}{-}   & \multicolumn{1}{c|}{-}    & \multicolumn{1}{c|}{-} & \multicolumn{1}{c|}{-}   & \multicolumn{1}{c|}{-}    & \multicolumn{1}{c|}{-} \\ \hline
NeuroLKH (2021) \cite{xin_neurolkh_2021}    			&L2P-O/LKH           	       & \multicolumn{1}{c|}{-}     & \multicolumn{1}{c|}{-}  & \multicolumn{1}{c|}{-} & \multicolumn{1}{c|}{7.765} & \multicolumn{1}{c|}{}   & \multicolumn{1}{c|}{} & \multicolumn{1}{c|}{-}     & \multicolumn{1}{c|}{-}  & \multicolumn{1}{c|}{-} & \multicolumn{1}{c|}{15.575} & \multicolumn{1}{c|}{-} & \multicolumn{1}{c|}{-} \\ \hline
GNN (2022) \cite{hudson_graph_2022} $\dagger$ 	&L2P-O/GLS             	       & \multicolumn{1}{c|}{-}     & \multicolumn{1}{c|}{0.00\%} & \multicolumn{1}{c|}{2.8h} & \multicolumn{1}{c|}{-}     & \multicolumn{1}{c|}{0.58\%} & \multicolumn{1}{c|}{2.8h} & \multicolumn{1}{c|}{-}     & \multicolumn{1}{c|}{-}    & \multicolumn{1}{c|}{-} & \multicolumn{1}{c|}{-}     & \multicolumn{1}{c|}{-}    & \multicolumn{1}{c|}{-} \\ \hline
DPDP (2022) \cite{kool_deep_2022} $\dagger$ 	&L2P-O/DP	              	       & \multicolumn{1}{c|}{-}     & \multicolumn{1}{c|}{-}    & \multicolumn{1}{c|}{-} & \multicolumn{1}{c|}{7.765} & \multicolumn{1}{c|}{0.00\%} & \multicolumn{1}{c|}{1.9h} & \multicolumn{1}{c|}{-}     & \multicolumn{1}{c|}{-}    & \multicolumn{1}{c|}{-} & \multicolumn{1}{c|}{15.627} & \multicolumn{1}{c|}{0.41\%}    & \multicolumn{1}{c|}{1.2d} \\ \hline
DIMES (2022) \cite{qiu_dimes_2022} $\dagger$  	 &L2P-O/MCTS     	       & \multicolumn{1}{c|}{-}     & \multicolumn{1}{c|}{-}    & \multicolumn{1}{c|}{-} & \multicolumn{1}{c|}{7.765} & \multicolumn{1}{c|}{0.01\%} & \multicolumn{1}{c|}{-} & \multicolumn{1}{c|}{-}     & \multicolumn{1}{c|}{-}    & \multicolumn{1}{c|}{-} & \multicolumn{1}{c|}{-}     & \multicolumn{1}{c|}{-}    & \multicolumn{1}{c|}{-} \\ \hline
UTSP (2023) \cite{min2023unsupervised}  	   &L2P-O/BFLS                       & \multicolumn{1}{c|}{5.696} & \multicolumn{1}{c|}{0.00\%}   & \multicolumn{1}{c|}{-} & \multicolumn{1}{c|}{7.765} & \multicolumn{1}{c|}{0.00\%} & \multicolumn{1}{c|}{-} & \multicolumn{1}{c|}{-}     & \multicolumn{1}{c|}{-}    & \multicolumn{1}{c|}{-} & \multicolumn{1}{c|}{-}     & \multicolumn{1}{c|}{-}    & \multicolumn{1}{c|}{-} \\ \hline
DIFUSCO (2023) \cite{sun_difusco_2023} $\dagger$  	 &L2P-O/2-opt        	       & \multicolumn{1}{c|}{5.696} & \multicolumn{1}{c|}{0.01\%} & \multicolumn{1}{c|}{5.8h} & \multicolumn{1}{c|}{7.766} & \multicolumn{1}{c|}{0.02\%} & \multicolumn{1}{c|}{21.7h} & \multicolumn{1}{c|}{-}     & \multicolumn{1}{c|}{-}    & \multicolumn{1}{c|}{-} & \multicolumn{1}{c|}{-}     & \multicolumn{1}{c|}{-}    & \multicolumn{1}{c|}{-} \\ \hline
T2TCO (2023) \cite{li_distribution_2023}    			&L2P-O/2-opt           	       & \multicolumn{1}{c|}{5.696}  & \multicolumn{1}{c|}{0.02\%}    & \multicolumn{1}{c|}{-} & \multicolumn{1}{c|}{7.767}  & \multicolumn{1}{c|}{0.06\%}   & \multicolumn{1}{c|}{-} & \multicolumn{1}{c|}{-}     & \multicolumn{1}{c|}{-}    & \multicolumn{1}{c|}{-} & \multicolumn{1}{c|}{-}     & \multicolumn{1}{c|}{-}    & \multicolumn{1}{c|}{-} \\ \hline
DeepACO (2023) \cite{ye_deepaco_2023}    				 &L2P-O/ACO       	       & \multicolumn{1}{c|}{-}     & \multicolumn{1}{c|}{-}    & \multicolumn{1}{c|}{-} & \multicolumn{1}{c|}{7.767}     & \multicolumn{1}{c|}{-}    & \multicolumn{1}{c|}{-} & \multicolumn{1}{c|}{-}     & \multicolumn{1}{c|}{-}    & \multicolumn{1}{c|}{-} & \multicolumn{1}{c|}{15.770}     & \multicolumn{1}{c|}{-}    & \multicolumn{1}{c|}{-} \\ \hline
EOH (2024) \cite{liu2024example}    				 &L2P-M/GLS       	       & \multicolumn{1}{c|}{-}     & \multicolumn{1}{c|}{0.00\%}    & \multicolumn{1}{c|}{-} & \multicolumn{1}{c|}{-}     & \multicolumn{1}{c|}{0.03\%}    & \multicolumn{1}{c|}{-} & \multicolumn{1}{c|}{-}     & \multicolumn{1}{c|}{-}    & \multicolumn{1}{c|}{-} & \multicolumn{1}{c|}{-}     & \multicolumn{1}{c|}{-}    & \multicolumn{1}{c|}{-} \\ \hline
ReEvo (2024) \cite{ye2024reevo}    				 &L2P-M/GLS       	       & \multicolumn{1}{c|}{-}     & \multicolumn{1}{c|}{0.00\%}    & \multicolumn{1}{c|}{-} & \multicolumn{1}{c|}{-}     & \multicolumn{1}{c|}{0.00\%}    & \multicolumn{1}{c|}{-} & \multicolumn{1}{c|}{-}     & \multicolumn{1}{c|}{-}    & \multicolumn{1}{c|}{-} & \multicolumn{1}{c|}{-}     & \multicolumn{1}{c|}{-}    & \multicolumn{1}{c|}{-} \\ \hline
Hercules (2025) \cite{wu2025}    				 &L2P-M/GLS      	       & \multicolumn{1}{c|}{-}     & \multicolumn{1}{c|}{0.00\%}    & \multicolumn{1}{c|}{-} & \multicolumn{1}{c|}{-}     & \multicolumn{1}{c|}{0.00\%}    & \multicolumn{1}{c|}{-} & \multicolumn{1}{c|}{-}     & \multicolumn{1}{c|}{-}    & \multicolumn{1}{c|}{-} & \multicolumn{1}{c|}{-}     & \multicolumn{1}{c|}{-}    & \multicolumn{1}{c|}{-} \\ \hline

\end{tabular}}
\begin{tablenotes}
\item[] Note: Symbol $\dagger$ denotes the results are taken from \cite{ma_learning_2023}, while the rest are taken from the given references. Abbreviations GLS and BFLS denote Guided Local Search \cite{voudouris1999guided} and Best-first Local Search \cite{xie2014adding}, respectively. For all experiments on TSP and CVRP, ``Obj.", ``Gap", and ``Time" denote the average objective values, average gaps, and the total run time on the corresponding test dataset with 10k instances, respectively. The gaps are computed relative to the exact algorithm Concorde \cite{applegate_the_2007} for TSP and the SOTA OR algorithm HGS \cite{vidal_hybird_2022} for CVRP. 
\end{tablenotes}
\end{table*}

To strive for high-quality solutions, Hottung et al. \cite{hottung_learning_2021} exploited an NN to predict the latent search space with a smooth landscape, then employed OR algorithms (e.g., Differential Evolution (DE) \cite{das_differential_2009}) to search in the predicted space. Recently, Liu et al. \cite{liu2023algorithm} proposed a framework based on Large Language Models (LLMs), to automatically generate multiple heuristic algorithms for producing high-quality solutions for VRPs. Similar to Evolutionary Computation (EC), the proposed LLM framework encompasses initialization, selection, crossover, and mutation steps. However, unlike EC, where each individual represents a solution of a VRP instance, in the LLM framework, each individual embodies a heuristic algorithm \cite{liu2023algorithm}. In addition, during each iteration, crossover and mutation operators involve employing prompts (comprising task descriptions, parent individuals, and generation instructions) to guide LLMs in generating multiple heuristic algorithms. Following \cite{liu2023algorithm}, Ye et al. \cite{ye2024reevo} further proposed two innovative prompt methods for crossover and mutation operators, respectively. The short-term prompt is employed for the crossover operator, compelling LLMs to analyze the differences in performance between the two parents as search directions. Conversely, the long-term prompt is utilized for the mutation operator, prompting LLMs to assimilate insights from historical individuals. To enable LLMs to provide specific search directions, Wu et al. \cite{wu2025} proposed the Core Abstraction Prompting method, which abstracts the core components from elite heuristics and incorporates them as prior knowledge in short- and long-term prompts. Future research could further encourage LLMs to develop more sophisticated heuristic algorithms that may yield superior performance. Moreover, a prior study \cite{wang2024distance} suggested that heuristics can enhance the performance of NCO solvers. This also represents a promising direction for improving the generalization ability of NCO solvers across different distributions and scales by deriving tailored heuristics through LLMs.

In conclusion, akin to L2P-O solvers, L2P-M solvers are extensively employed to expedite OR algorithm search processes or improve solution quality. However, generally speaking, L2P-M solvers tend to be more time-consuming in similar situations due to their multiple-prediction nature.
\subsection{Performance of L2P-O and L2P-M Solvers}
\label{sec5.3}

Table~\ref{table4} presents the performance of existing L2P-O and L2P-M solvers on small-scale TSP and CVRP. The results demonstrate that by leveraging the advantages of the conventional OR or search algorithms, most L2P-O and \mbox{L2P-M} solvers exhibit commendable performance on small-scale VRP instances, such as the 50-node TSP. Furthermore, these \mbox{L2P-O} and L2P-M solvers exhibit proficiency in handling large-scale TSP instances, outperforming L2C and L2I solvers (see Section~\ref{sec6.2} for more details). However, other than DPDP \cite{kool_deep_2022} and DeepACO \cite{ye_deepaco_2023}, the remaining  \mbox{heatmap-based} \mbox{L2P-O} solvers, including GCN \cite{joshi_efficient_2019}, UTSP \cite{min2023unsupervised}, Att-GCN \cite{fu_generalize_2021}, DIMES \cite{qiu_dimes_2022}, DIFUSCO \cite{sun_difusco_2023}, and T2TCO \cite{li_distribution_2023}, encounter difficulties when addressing CVRP. This difficulty arises from the additional constraints imposed in CVRP, i.e., the varying goods demand for each node and a vehicle capacity limit. The intricate constraints of VRP variants cannot be adequately modeled solely through the predicted heatmaps and search algorithms \cite{ma_learning_2023}. In future research, the exploration of integrating more robust OR algorithms is warranted to address the diverse constraints inherent in various VRP variants and improve the overall performance of heatmap-based solvers.

\section{Inadequacies, Efforts, and Perspectives of Existing NCO Solvers }
\label{sec6}
Despite the high-level performance achieved by existing NCO solvers in terms of the quality of solutions and time complexity, they still face certain inadequacies when compared to the conventional OR algorithms \cite{liu_how_2023, fran2019evaluate, caramanis_optimizing_2023, Sun_Revisiting_2023, gao2024generalizable}. Sections~\ref{sec6.1} to \ref{sec6.4} introduce four specific inadequacies of existing NCO solvers, discuss existing efforts to solve these inadequacies, identify open inadequacies, and propose additional promising directions to address them.

\subsection{Inadequacies of Generalization Capability}
\label{sec6.1}
The prevailing inadequacy among the majority of existing NCO solvers is their limited capacity for generalization across diverse data distribution patterns \cite{bdeir2022attention, sultana_learning_2022, jiang2023ensemblebased, Wang_asp_2024,liu2024prompt, Zhou_Collaboration}. For example, solvers trained with instances following the uniform distribution (i.e., $\forall v_i\in\bm{V}, v_{i}\sim\mathcal{U}([0,1]^2)$) are less likely capable of handling instances conforming to the normal distribution (i.e., $\forall v_i\in\bm{V}, v_{i}\sim\mathcal{N}([0,1]^2)$) \cite{bi_learning_2022}. This inadequacy impedes the application of NCO solvers in real-world problems \cite{zhang_review_2023}. 

To address the inadequacy of impaired generalization ability, certain L2C solvers leveraged more diverse instances as training sets, compelling solvers to learn from varying types of distribution \cite{joshi_learning_2021, jiang_learning_2022, zhang_learning_2022, geisler_generalization_2022, jiang_multi_2023}. For example, Jiang et al. \cite{jiang_learning_2022} collectively optimized the parameters of solvers using instances from different distribution patterns. Meanwhile, Zhang et al. \cite{zhang_learning_2022} incorporated Curriculum Learning into L2C solvers, which utilizes instances with diverse distribution patterns to progressively train solvers. In contrast to the aforementioned approaches that focused on the diversity of instance distribution patterns, Bi et al. \cite{bi_learning_2022} tackled the inadequacy of generalization at the model level. Specifically, they incorporated the knowledge distillation method into NCO, utilizing multiple teacher models trained on different types of distribution to guide a generalized student model capable of performing effectively across varying distribution patterns.

In addition, NCO solvers exhibit limited generalization capability across different problem scales simultaneously \cite{joshi_learning_2021}. For example, those trained on trivially small scales struggle to generalize to larger instances. While the approaches introduced in the previous paragraph solely focus on distribution, certain studies \cite{manchanda_generalization_2023, zhou_towards_2023} simultaneously addressed generalization concerning both the problem scale and distribution. Specifically, Manchanda et al. \cite{manchanda_generalization_2023} formalized solving VRP instances over a given instance distribution and scale as a separate learning task. Subsequently, they utilized Meta Learning to learn a model capable of adapting to various tasks. Following \cite{manchanda_generalization_2023}, Zhou et al. \cite{zhou_towards_2023} replaced the random task sampling method used in \cite{manchanda_generalization_2023} with a hierarchical scheduler to adaptively select different types of task. In addition, to mitigate the computing overhead, they introduced a second-order derivative approximation method within the Meta Learning framework. This approach entails utilizing second-order derivatives primarily in the early stages of training, and subsequently relying exclusively on their first-order counterparts. 

While considerable advancement has been made in enhancing the generalization of NCO solvers, it is crucial to acknowledge that these generalization methods require more hyperparameters and efforts on data synthesis, resulting in a significant overhead in computing resources.  We propose employing more lightweight NNs to improve generalization capability. For example, Xiao et al. \cite{xiao2024improving} proposed a plug-and-play entropy-based scaling factor to adjust attention scores, thereby enhancing the generalization capability of solver while requiring fewer computing resources. In addition, these solvers fall under the purview of domain adaptation for addressing the out-of-distribution problem \cite{liu2021towards}. Specifically, these solvers focus on adapting to the instances from known target distributions. However, they exhibit inadequacies in effectively handling instances from unknown distribution patterns in real-world problems (e.g., TSPLib \cite{reinelt1991tsplib} and CVRPLib \cite{queiroga2022}). We suggest incorporating Invariant Learning into NCO solvers, directing their attention to data features unaffected by environmental variations, thus enhancing their generalization capability. Recent studies \cite{fang2024invit, xiao2025,gao2024generalizable} emphasized the role of nearest-neighbor relationships among nodes in capturing local invariance across problems of varying scales and distributions.

Currently, certain solvers achieve high-level performance on synthetic TSP datasets across different distributions and scales. For instance, the SOTA solver GELD \cite{xiao2025} attains gap values of 2.38\%, 3.44\%, 4.7\%, and 2.81\% on 10,000-node TSP instances with uniform, clustered, explosion, and implosion distributions, respectively. However, on real-world instances of the same scale, such as those in TSPLib\footnote{URL: http://comopt.ifi.uni-heidelberg.de/software/TSPLIB95/} and National TSPs\footnote{URL: https://www.math.uwaterloo.ca/tsp/world/countries.html}, GELD's gap value remains high at 7\% due to differences between the real-world distributions and the training distributions. This highlights the need for further improvement of NCO solvers on real-world instances. In addition, drawing from the emphasis on enhancing the generalization capacity of L2C solvers in prior studies, attention to other solver categories, such as L2I, L2P-O, and L2P-M solvers, has been relatively scarce. We posit that with the help of OR algorithms, L2P-O and L2P-M solvers could potentially demonstrate superior generalization capabilities.

\subsection{Inadequacies of Solving Large-scale VRPs}
\label{sec6.2}

In addition to the inadequacy posed by poor generalization across different problem distributions and scales, many existing solvers encounter difficulties when handling large-scale VRPs in real-time \cite{ Zhang_neural_2023}. This is particularly notable for L2C and L2I solvers \cite{ drakulic2023bqnco, chen4679437extnco}, despite their demonstrated ability to outperform or be comparable to the conventional OR algorithms with up to 100 nodes \cite{kwon_pomo_2020}. The underlying reason lies in the quadratic growth of time complexity and memory space associated with NNs in most solvers as the problem scale increases \cite{luo_neural_2023}. One prominent example is the self-attention mechanism, a widely adopted NN in NCO solvers. Self-attention mechanism is defined based on query $Q$, key $K$, value $V$, where $Q, K, V \in \mathbb{R}^{N \times d}$, with $N$ and $d$ denoting the problem scale and predefined dimension, respectively. The embedding $h_i$ for the $i$th node is computed using the following Softmax function:
\begin{equation}
\label{eq17}
     h_i = \sum\nolimits_j \frac{\exp{(q_i k_j^\top /\sqrt{d})}}{{\sum_l{\exp{(q_i k_l^\top /\sqrt{d})}}}}v_j,
\end{equation}
where $\exp(\cdot)$ and $\top$ denote exponential and transpose functions, respectively. Vectors $q_i$, $k_i$, and $v_i$ denote the query, key, and value of the $i$th node, respectively. According to (\ref{eq17}), it is evident that each instance involves a dot product computation with a complexity of $\mathcal{O}(N^2d)$ and a memory usage complexity of $\mathcal{O}(N^2)$ \cite{vaswani_attention_2017}. When the problem scale $N$ increases to $\lambda N$, where $\lambda$ denotes an integer greater than one, the complexity of dot product computation and memory usage grow to $\mathcal{O}(\lambda^2 N^2 d)$ and $\mathcal{O}(\lambda^2 N^2)$, respectively. Consequently, as the problem scale expands, solvers employing the self-attention mechanism are prone to memory depletion during training, hampering convergence \cite{luo_neural_2023}. Similarly, various other NNs experience problem scale-related effects to varying degrees. 
\begin{table*}[!t]
\centering

\caption{\label{table5} Performance of various NCO solves on large-scale TSP and CVRP instances}
	\scalebox{0.7}{
\begin{tabular}{|c|c|ccc|ccc|ccc|cc|cc|}
\hline
\multirow{2}{*}{Methods} & \multirow{2}{*}{Type}  & \multicolumn{3}{c|}{TSP ($N=500$)}                                                  & \multicolumn{3}{c|}{TSP ($N=1,000$)}                                                  & \multicolumn{3}{c|}{TSP ($N=10,000$)}   &   \multicolumn{2}{c|}{CVRP ($N=1,000$)}   &   \multicolumn{2}{c|}{CVRP ($N=2,000$)}                                                                                \\ \cline{3-15} 
                      &  & 
\multicolumn{1}{c|}{Obj.} & \multicolumn{1}{c|}{Gap} & Time                       & \multicolumn{1}{c|}{Obj.} & \multicolumn{1}{c|}{Gap} & Time                  & \multicolumn{1}{c|}{Obj.} & \multicolumn{1}{c|}{Gap} & Time         & 
\multicolumn{1}{c|}{Obj.} & Time & 
\multicolumn{1}{c|}{Obj.} & Time     \\ \hline
LKH3 (2021) \cite{helsgaun2017extension} $\ddagger$ &  OR		              	      &  \multicolumn{1}{c|}{16.55}  & \multicolumn{1}{c|}{-} & \multicolumn{1}{c|}{5.5m} & \multicolumn{1}{c|}{23.12}  & \multicolumn{1}{c|}{-} & \multicolumn{1}{c|}{24m} & \multicolumn{1}{c|}{71.77}  & \multicolumn{1}{c|}{-} & \multicolumn{1}{c|}{13h} & 
\multicolumn{1}{c|}{46.40}  & \multicolumn{1}{c|}{6.2s} & 
\multicolumn{1}{c|}{64.90} & \multicolumn{1}{c|}{20s} \\ \hline
Att-GCN (2021) \cite{fu_generalize_2021} $\ddagger$ &  L2P-O/D\&C/MCTS		          	      &  \multicolumn{1}{c|}{16.96} & \multicolumn{1}{c|}{2.48\%} & \multicolumn{1}{c|}{2.9m} & \multicolumn{1}{c|}{23.86} & \multicolumn{1}{c|}{3.20\%} & \multicolumn{1}{c|}{6.1m} & \multicolumn{1}{c|}{75.73} & \multicolumn{1}{c|}{5.50\%} & \multicolumn{1}{c|}{13.1m} & 
\multicolumn{1}{c|}{-}  & \multicolumn{1}{c|}{-} & 
\multicolumn{1}{c|}{-} & \multicolumn{1}{c|}{-}  \\ \hline
L2D (2021) \cite{li_learning_2021} $\ddagger$ &  L2P-M/D\&C/LKH3		              	      &  \multicolumn{1}{c|}{-} & \multicolumn{1}{c|}{-} & \multicolumn{1}{c|}{-} & \multicolumn{1}{c|}{-} & \multicolumn{1}{c|}{-} & \multicolumn{1}{c|}{-} & \multicolumn{1}{c|}{-} & \multicolumn{1}{c|}{-} & \multicolumn{1}{c|}{-} & 
\multicolumn{1}{c|}{46.30}  & \multicolumn{1}{c|}{1.5s} & 
\multicolumn{1}{c|}{65.20} & \multicolumn{1}{c|}{38s}  \\ \hline
DIMES (2022) \cite{qiu_dimes_2022} $\ddagger$ & 		L2P-O/MCTS	           	       &  \multicolumn{1}{c|}{17.01}  & \multicolumn{1}{c|}{2.78\%} & \multicolumn{1}{c|}{3.1m} & \multicolumn{1}{c|}{23.86}  & \multicolumn{1}{c|}{3.20\%} & \multicolumn{1}{c|}{3.6m} & \multicolumn{1}{c|}{76.02}  & \multicolumn{1}{c|}{5.90\%} & \multicolumn{1}{c|}{33.7m} & 
\multicolumn{1}{c|}{-}  & \multicolumn{1}{c|}{-} & 
\multicolumn{1}{c|}{-} & \multicolumn{1}{c|}{-}\\ \hline
TSPformer (2023) \cite{yang_memory_2023} $\ddagger$ &  	L2C		            	       &  \multicolumn{1}{c|}{17.57} & \multicolumn{1}{c|}{5.97\%} & \multicolumn{1}{c|}{3.1m} & \multicolumn{1}{c|}{27.02} & \multicolumn{1}{c|}{16.90\%} & \multicolumn{1}{c|}{5.0m} & \multicolumn{1}{c|}{-}     & \multicolumn{1}{c|}{-}    & \multicolumn{1}{c|}{-} & 
\multicolumn{1}{c|}{-}  & \multicolumn{1}{c|}{-} & 
\multicolumn{1}{c|}{-} & \multicolumn{1}{c|}{-} \\ \hline
Pointerformer (2023) \cite{jin_pointerformer_2023} $\ddagger$ &  L2C     	       &
 \multicolumn{1}{c|}{17.14} & \multicolumn{1}{c|}{3.56\%} & \multicolumn{1}{c|}{1.0m} & \multicolumn{1}{c|}{24.80} & \multicolumn{1}{c|}{7.30\%} & \multicolumn{1}{c|}{6.5m} &  \multicolumn{1}{c|}{-}     & \multicolumn{1}{c|}{-}    & \multicolumn{1}{c|}{-} & 
\multicolumn{1}{c|}{-}  & \multicolumn{1}{c|}{-} & 
\multicolumn{1}{c|}{-} & \multicolumn{1}{c|}{-} \\ \hline
H-TSP (2023) \cite{pan_h-tsp_2023}   $\ddagger$ 		&	L2C/D\&C 	              	       &  \multicolumn{1}{c|}{-}     & \multicolumn{1}{c|}{-}    & \multicolumn{1}{c|}{-} & \multicolumn{1}{c|}{24.65}  & \multicolumn{1}{c|}{6.62\%}  & \multicolumn{1}{c|}{47s} & \multicolumn{1}{c|}{77.75}  & \multicolumn{1}{c|}{7.32\%}  & \multicolumn{1}{c|}{48s} & 
\multicolumn{1}{c|}{-}  & \multicolumn{1}{c|}{-} & 
\multicolumn{1}{c|}{-} & \multicolumn{1}{c|}{-} \\ \hline
TAM-AM (2023) \cite{hou_generalize_2023}   $\ddagger$ 		&	L2C/D\&C/	       &  \multicolumn{1}{c|}{-}     & \multicolumn{1}{c|}{-}    & \multicolumn{1}{c|}{-} & \multicolumn{1}{c|}{-}  & \multicolumn{1}{c|}{-}  & \multicolumn{1}{c|}{-} & \multicolumn{1}{c|}{-}  & \multicolumn{1}{c|}{-}  & \multicolumn{1}{c|}{-} & 
\multicolumn{1}{c|}{50.06}  & \multicolumn{1}{c|}{0.8s} & 
\multicolumn{1}{c|}{74.31} & \multicolumn{1}{c|}{2.2s} \\ \hline
TAM-LKH3 (2023) \cite{hou_generalize_2023}   $\ddagger$ 		&	L2P-O/D\&C/LKH3	       &  \multicolumn{1}{c|}{-}     & \multicolumn{1}{c|}{-}    & \multicolumn{1}{c|}{-} & \multicolumn{1}{c|}{-}  & \multicolumn{1}{c|}{-}  & \multicolumn{1}{c|}{-} & \multicolumn{1}{c|}{-}  & \multicolumn{1}{c|}{-}  & \multicolumn{1}{c|}{-} & 
\multicolumn{1}{c|}{46.30}  & \multicolumn{1}{c|}{1.8s} & 
\multicolumn{1}{c|}{64.80} & \multicolumn{1}{c|}{5.6s} \\ \hline
SO-DR (2023) \cite{cheng_select_2023}    		&L2P-M/D\&C/LKH3               	       &  \multicolumn{1}{c|}{16.94}  & \multicolumn{1}{c|}{2.40\%}    & \multicolumn{1}{c|}{-} & \multicolumn{1}{c|}{23.77}  & \multicolumn{1}{c|}{2.80\%}    & \multicolumn{1}{c|}{-} & \multicolumn{1}{c|}{74.30}  & \multicolumn{1}{c|}{3.52\%}    & \multicolumn{1}{c|}{-} & 
\multicolumn{1}{c|}{-}  & \multicolumn{1}{c|}{-} & 
\multicolumn{1}{c|}{-} & \multicolumn{1}{c|}{-} \\ \hline
UTSP (2023) \cite{min2023unsupervised}    		&	L2P-O/BFLS	              	       &  \multicolumn{1}{c|}{16.68}  & \multicolumn{1}{c|}{0.84\%}  & \multicolumn{1}{c|}{-} & \multicolumn{1}{c|}{23.39}  & \multicolumn{1}{c|}{1.17\%}  & \multicolumn{1}{c|}{-} & \multicolumn{1}{c|}{-}     & \multicolumn{1}{c|}{-}    & \multicolumn{1}{c|}{-} & 
\multicolumn{1}{c|}{-}  & \multicolumn{1}{c|}{-} & 
\multicolumn{1}{c|}{-} & \multicolumn{1}{c|}{-} \\ \hline
LEHD (2023) \cite{luo_neural_2023}    	&	L2C/RRC              	       &  \multicolumn{1}{c|}{-}     & \multicolumn{1}{c|}{0.48\%}    & \multicolumn{1}{c|}{-} & \multicolumn{1}{c|}{-}     & \multicolumn{1}{c|}{1.22\%}    & \multicolumn{1}{c|}{-} & \multicolumn{1}{c|}{-}     & \multicolumn{1}{c|}{-}    & \multicolumn{1}{c|}{-} & 
\multicolumn{1}{c|}{-}  & \multicolumn{1}{c|}{2.37\%} & 
\multicolumn{1}{c|}{-} & \multicolumn{1}{c|}{-} \\ \hline
DIFUSCO (2023) \cite{sun_difusco_2023}    &	L2P-O/2-opt		              	       &  \multicolumn{1}{c|}{16.63} & \multicolumn{1}{c|}{0.46\%}    & \multicolumn{1}{c|}{-} & \multicolumn{1}{c|}{23.39} & \multicolumn{1}{c|}{1.17\%}    & \multicolumn{1}{c|}{-} & \multicolumn{1}{c|}{73.62} & \multicolumn{1}{c|}{2.58\%}    & \multicolumn{1}{c|}{-} & 
\multicolumn{1}{c|}{-}  & \multicolumn{1}{c|}{-} & 
\multicolumn{1}{c|}{-} & \multicolumn{1}{c|}{-} \\ \hline
T2TCO (2023) \cite{li_distribution_2023}   & 		L2P-O/2-opt	              	       &  \multicolumn{1}{c|}{16.61} & \multicolumn{1}{c|}{0.37\%}    & \multicolumn{1}{c|}{-} & \multicolumn{1}{c|}{23.30} & \multicolumn{1}{c|}{0.78\%}    & \multicolumn{1}{c|}{-} & \multicolumn{1}{c|}{-}     & \multicolumn{1}{c|}{-}    & \multicolumn{1}{c|}{-} & 
\multicolumn{1}{c|}{-}  & \multicolumn{1}{c|}{-} & 
\multicolumn{1}{c|}{-} & \multicolumn{1}{c|}{-}  \\ \hline
DeepACO (2023) \cite{ye_deepaco_2023} $\ddagger$ & L2P-O/ACO	              	       &  \multicolumn{1}{c|}{16.94} & \multicolumn{1}{c|}{2.36\%} & \multicolumn{1}{c|}{4.3m} & \multicolumn{1}{c|}{23.85} & \multicolumn{1}{c|}{3.16\%} & \multicolumn{1}{c|}{1.1h} & \multicolumn{1}{c|}{-} & \multicolumn{1}{c|}{-} & \multicolumn{1}{c|}{-} & 
\multicolumn{1}{c|}{-}  & \multicolumn{1}{c|}{-} & 
\multicolumn{1}{c|}{-} & \multicolumn{1}{c|}{-}\\ \hline
GLOP (2024) \cite{ye_glop_2024} $\ddagger$ & 	L2I/D\&C              	       &  \multicolumn{1}{c|}{16.91} & \multicolumn{1}{c|}{1.99\%} & \multicolumn{1}{c|}{1.5m} & \multicolumn{1}{c|}{23.84} & \multicolumn{1}{c|}{3.11\%} & \multicolumn{1}{c|}{3.0m} & \multicolumn{1}{c|}{75.29} & \multicolumn{1}{c|}{4.90\%} & \multicolumn{1}{c|}{1.8m} & 
\multicolumn{1}{c|}{-} & \multicolumn{1}{c|}{-} & 
\multicolumn{1}{c|}{-} & \multicolumn{1}{c|}{-}\\ \hline
GLOP-G (2024) \cite{ye_glop_2024} $\ddagger$ & 	L2P-O/D\&C/LKH3                 &  \multicolumn{1}{c|}{-} & \multicolumn{1}{c|}{-} & \multicolumn{1}{c|}{-} & \multicolumn{1}{c|}{-} & \multicolumn{1}{c|}{-} & \multicolumn{1}{c|}{-} & \multicolumn{1}{c|}{-} & \multicolumn{1}{c|}{-} & \multicolumn{1}{c|}{-} & 
\multicolumn{1}{c|}{45.90} & \multicolumn{1}{c|}{1.1s} & 
\multicolumn{1}{c|}{63.00} & \multicolumn{1}{c|}{1.5s}\\ \hline
UDC (2024) \cite{zheng2024udc}  & 	L2C/D\&C       &  \multicolumn{1}{c|}{-} & \multicolumn{1}{c|}{1.58\%} & \multicolumn{1}{c|}{-} & \multicolumn{1}{c|}{-} & \multicolumn{1}{c|}{1.78\%} & \multicolumn{1}{c|}{-} & \multicolumn{1}{c|}{-} & \multicolumn{1}{c|}{-} & \multicolumn{1}{c|}{-} & 
\multicolumn{1}{c|}{43.00} & \multicolumn{1}{c|}{-} & 
\multicolumn{1}{c|}{60.01} & \multicolumn{1}{c|}{-}\\ \hline
DEITSP (2025) \cite{mingzhao}  & 	L2P-O/2-opt         &  \multicolumn{1}{c|}{-} & \multicolumn{1}{c|}{2.15\%} & \multicolumn{1}{c|}{-} & \multicolumn{1}{c|}{-} & \multicolumn{1}{c|}{3.68\%} & \multicolumn{1}{c|}{-} & \multicolumn{1}{c|}{-} & \multicolumn{1}{c|}{-} & \multicolumn{1}{c|}{-} & 
\multicolumn{1}{c|}{-} & \multicolumn{1}{c|}{-} & 
\multicolumn{1}{c|}{-} & \multicolumn{1}{c|}{-}\\ \hline
GELD (2025) \cite{xiao2025}  & 	L2C       &  \multicolumn{1}{c|}{-} & \multicolumn{1}{c|}{0.52\%} & \multicolumn{1}{c|}{-} & \multicolumn{1}{c|}{-} & \multicolumn{1}{c|}{0.58\%} & \multicolumn{1}{c|}{-} & \multicolumn{1}{c|}{-} & \multicolumn{1}{c|}{2.38\%} & \multicolumn{1}{c|}{-} & 
\multicolumn{1}{c|}{-} & \multicolumn{1}{c|}{-} & 
\multicolumn{1}{c|}{-} & \multicolumn{1}{c|}{-}\\ \hline
\end{tabular}}
\begin{tablenotes}
\item[] Note: Symbol $\ddagger$ denotes the results are taken from \cite{ye_glop_2024}, and the rest are taken from the given references. For all experiments on TSP and CVRP, the terms ``Obj." and ``Gap" denote the average objective values and average gaps, respectively. The gaps are computed in comparison to the heuristic solver LKH3 \cite{helsgaun2017extension} for both TSP and CVRP. In addition, following the prior study \cite{ye_glop_2024}, ``Time" denotes the overall and average running times on the corresponding TSP and CVRP test datasets, respectively. The TSP test datasets for 500-node and 1,000-node each consist of 128 instances, while the 10,000-node TSP test dataset includes 16 instances. The CVRP test datasets for 1,000-node and 2,000-node each consist of 100 instances.
\end{tablenotes}
\end{table*}
To address this issue, certain methods have embraced the D\&C manner to decompose large-scale problems into multiple small-scale sub-problems \cite{pan_h-tsp_2023, fu_generalize_2021, li_learning_2021, ye_glop_2024, hou_generalize_2023}. Subsequently, these methods employ diverse solvers to construct or improve solutions for all decomposed sub-problems, followed by the integration of all sub-problems. For example, Pan et al. \cite{pan_h-tsp_2023} proposed H-TSP, which utilizes an upper-level partition model and a lower-level L2C solver for partitioning and solving sub-problems, respectively. Similarly, Hou et al. \cite{hou_generalize_2023} exploited a Transformer to divide a problem into multiple sub-problems and incorporated LKH3 to construct solutions for each sub-problem. For these D\&C solvers, identifying the optimal decomposition strategy for large-scale VRPs stands as a crucial factor \cite{zheng2024udc,ye_glop_2024}.

While pursuing the same objective, other NCO solvers \cite{min2023unsupervised, jin_pointerformer_2023, yang_memory_2023,zhou2024instance} diverge from the D\&C manner in addressing the time and memory expenses associated with solving large-scale VRPs. Specifically, they adopt lighter NNs or innovative training paradigms. For example, Yang et al. \cite{yang_memory_2023} replaced scaled dot-product attention with sampled scaled dot-production attention (see Section~\ref{sec3.1} for more details). Xiao et al. \cite{xiao2025} proposed the Region Average Linear Attention mechanism, which partitions nodes into regions to facilitate efficient global information exchange, achieving a computation complexity $\mathcal{O}(Nd)$. In addition, to optimize memory usage, Luo et al. \cite{luo_neural_2023} adopted SL as the training paradigm to train an NN rather than RL. This choice stems from the fact that RL training necessitates generating the complete solution before computing the reward, implying a substantial demand for memory and computing resources. More recently, studies \cite{pirnay2024self} and \cite{luo2025boosting} exploited the high-quality solutions produced by the solver as labels, enabling self-supervised learning. Furthermore, in contrast to the original study \cite{joshi_efficient_2019}, which involves training a GCN using SL, Min et al. \cite{min2023unsupervised} introduced an innovative UL loss function to train GCN. By adopting this design, GCN utilizes approximately 10\% of the number of parameters compared to the SL training paradigm.

Table~\ref{table5} presents the performance of NCO solvers designed for solving large-scale TSP and CVRP instances. For TSP, comparing L2C solvers that abstain from employing the D\&C approach (TSPformer \cite{yang_memory_2023} and Pointerformer \cite{jin_pointerformer_2023}) with their counterparts utilizing this strategy (H-TSP \cite{pan_h-tsp_2023}) exposes a consistent performance gap ranging from 1\% to 10\%. However, even though H-TSP adopts the D\&C manner, its capability is limited to handling TSP instances with a maximum of 1,000 nodes. In contrast, L2P-O solvers outperform L2C solvers, especially the recently proposed heatmap-based solvers (e.g., UTSP \cite{min2023unsupervised}, DIFUSCO \cite{sun_difusco_2023}, T2TCO \cite{li_distribution_2023}, and Fast T2T \cite{fastt2t}) that can achieve high-level performance in real-time, with gaps falling within the 1\% range, when dealing with 500-node and 1,000-node TSP instances. Notably, on the challenging 10,000-node TSP, DIFUSCO \cite{sun_difusco_2023} achieved an impressive gap value of 2.58\%, credited to the training method exploiting the diffusion model. Specifically, DIFUSCO constantly adds Bernoulli noises to optimal solutions of instances and compels a GCN to denoise the added noises. Through this training method, GCN has the capability to predict high-quality heatmaps. Subsequently, MCTS is employed to construct (sub-)optimal solutions based on these high-quality heatmaps. To accelerate inference, Wang et al. \cite{mingzhao} proposed DEITSP, which optimizes multiple generation steps of DIFUSCO into a single step, transforming  Bernoulli noise into high-quality heatmaps through a specially designed bimodal graph Transformer for one-step prediction.

For CVRP, a VRP variant beyond TSP, heatmap-based solvers lack the generality to handle it efficiently \cite{ma_learning_2023, mingzhao}. Because the intricate capacity constraint of CVRP cannot be adequately modeled solely through the predicted heatmaps, as mentioned in Section~\ref{sec5.3}. We propose to employ more sophisticated OR algorithms (e.g., DP) for heatmap-based solvers, rather than relying solely on basic search algorithms. The incorporation of OR algorithms presents a viable direction to this problem. To solve large-scale CVRP instances, current NCO solvers generally adopt the D\&C manner. For example, Zheng et al. \cite{zheng2024udc} proposed a unified neural D\&C framework (UDC), which employs an Anisotropic Graph Neural Network for global node partitioning and subsequently adopts an L2C solver to solve all CVRP sub-problems, achieving performance that surpasses LKH3.

Although certain solvers have attained strong performance on synthetic large-scale TSP and CVRP instances, their effectiveness on other VRP variants still requires improvement. For instance, although UDC \cite{zheng2024udc} employs the same L2C solver (i.e., ICAM \cite{zhou2024instance}) for both CVRP and Open VRP (OVRP), its performance on the 2,000-node OVRP instances lags behind LKH3 by 8.71\%. For future work in this regard, we suggest utilizing more lightweight NNs (e.g., Mamba \cite{gu2023mamba}) as replacements for the widely adopted Transformer in L2C solvers. This design adjustment enables L2C solvers to solve VRP with constraints from a global perspective, thereby mitigating the distance cost associated with inadequate decomposition strategies in the D\&C manner. In addition, as mentioned in Section~\ref{sec6.1}, there is a performance gap between existing solvers on synthetic datasets and real-world ones, which is an inadequacy that warrants further investigation in the future.

\subsection{Inadequacies of Vehicle Routing Problem Variants}
\label{sec6.3}
As shown in Section~\ref{sec2.1}, most NCO solvers specialize in TSP and CVRP, exhibiting inadequacies when faced with VRP incorporating other constraints \cite{darvariu2024graph, luttmann2024neural,10485273}. Consequently, there arises a necessity to formulate solvers explicitly designed for VRP variants with intricate constraints in realistic scenarios \cite{tong_combinatorial_2023}. For example, dispatching vehicle fleets within the premise of an airport can be considered a multiple-fleet VRP \cite{Zhou_Learning_2023}, encompassing various constraints such as capacity, time window, precedence, etc. In this subsection, we investigate methodologies for customizing NCO solvers to address the constraints posed by MOVRP and DVRP, two prominent VRP variants entailing constraints in real-world scenarios.

To tackle the problem arising from the multiple objectives in MOVRP, certain studies \cite{Wu_2020_MODRLDAM, Li_deep_2021, Zhang_meta_2022, zhang_MODRL_2021} adopted the MOEA/D framework \cite{zhang_MOEA_2007}. This framework is designed to decompose a multi-objective problem into multiple distinct single-objective problems, and develops models corresponding to each single-objective problem, thereby addressing them individually. Finally, the Pareto set is constructed based on solutions of all single-objective problems. Furthermore, Lin et al. \cite{lin_pareto_2022} proposed an approach to approximate the Pareto set using a single preference-conditioned solver, instead of employing multiple solvers. This preference-conditioned solver takes preference vectors (each vector denotes a single-objective problem) as inputs and constructs all possible (sub-)optimal solutions on the Pareto front. In addition, certain studies have demonstrated the value of incorporating the imaging modality to address MOVRP. Chen et al. \cite{chen2025neural} proposed a graph-image bimodal fusion framework that enhances the performance of NCO by integrating graph and image information.

When dealing with DVRP, existing NCO solvers choose real-time embedding updates to meet the real-time requirements \cite{zhang_solving_2023, paul_multi_2022, sivagnanam_offline_2022, joe2020deep, Yu_online_2019, pan_deep_2023}, as mentioned in Section~\ref{sec3.1}. For example, Zhang et al. \cite{zhang_solving_2023} proposed an L2C solver that handles a dynamic node pool. During each decoding step, this node pool is adjusted by removing or adding nodes to simulate cancelled or new customer orders. Subsequently, they employed the updated node pool to update current embeddings. Moreover, in real-world scenarios, e.g., ride-hailing, NCO solvers encounter the dual challenge of dynamically updating nodes requiring visits while simultaneously adhering to time window constraints \cite{tong_combinatorial_2023, sivagnanam_offline_2022, joe2020deep}.

Recently, several studies \cite{liu2024multitask, wang2023efficient, lin2024crossproblem,mvmoe,li2024cada, berto2025route} integrated Multi-task Learning into the field of NCO, enabling solvers to adeptly tackle a range of VRP variants. For example, Liu et al. \cite{liu2024multitask} incorporated the concept of compositional zero-shot learning to propose MTL, which formulates VRPs as different combinations of a set of constraints (e.g., capacity) and exploits a shared network to encode these constraints. To train the solver, MTL employs a joint loss function comprising multiple sub-loss functions of equal weightage, each corresponding to a distinct VRP variant. Zhou et al. \cite{mvmoe} incorporated the concept of Mixture-of-Experts to further enhance the performance of MTL. However, learning multiple tasks simultaneously presents challenges not encountered in single-task learning, primarily due to potential conflicts between different task requirements \cite{Zhang_survey_2022, crawshaw2020multitask}. We advocate for alleviating conflicts by adjusting the weightage of all the sub-loss functions. For example, to better address this problem, Berto et al. \cite{berto2025route} proposed mixed batch training and multi-variant reward normalization. The former ensures that each training batch comprises VRP instances with diverse constraints, while the latter normalizes the reward for each VRP variant. In addition, Meta Learning offers a viable solution, which can discern relationships between tasks and dynamically adjust the weightages of their corresponding sub-loss functions. It is worth noting that certain studies \cite{drakulic2025goal, pan2025unico} have also explored using a versatile solver to address different COPs, such as VRP and JSP, a goal that poses greater challenges than designing a unified solver for VRP variants.

Currently, these versatile solvers are limited to handling VRPs of sizes up to 100 and cannot effectively handle instances from different distributions. Future research should focus on enhancing their performance on large-scale VRP instances and across various distributions, while progressively incorporating additional VRP variants. Once these powerful and versatile solvers have been developed to handle large-scale VRPs across diverse instance distributions and problem constraints, without the need of special designs and retraining, we anticipate the industry will be keen to adopt them in a wide range of applications.

\subsection{Inadequacies of Comparison Methods}
\label{sec6.4}

When comparing NCO solvers, fairness is only ensured if the gap is computed upon the same baseline, the test set sizes are equal, and the same hardware and parameters (e.g.,  the number of iterations) are utilized. However, the unfair comparison issue extends to a more significant issue when comparing NCO solvers with the conventional OR algorithms \cite{garmendia2022neural, accorsi_guidelines_2022, Berto_rl4co_2023}. NCO solvers inherently benefit from operating on GPU, whereas OR algorithms are traditionally executed on Central Processing Units (CPUs). Consequently, existing comparison methods unfairly favor NCO algorithms in terms of time, which  results in the fact that most NCO solvers are still rarely adopted by the OR community \cite{liu_how_2023}.

For a fairer comparison between NCO solvers and OR algorithms, Liu et al. \cite{liu_how_2023} simultaneously executed OR algorithms on multiple CPU threads to address multiple VRP instances. In addition, they introduced energy as a complementary metric in addition to performance gap and computation time. Specifically, energy denotes the electric power consumed by either NCO solvers or OR algorithms when solving test instances, recorded using the open-source PowerJoular tool \cite{noureddine_ie_2022}. In addition, Lu et al. \cite{lu_roco_2023} introduced the first robustness metric applicable to both OR algorithms and NCO solvers. This metric precisely measures the solver's cost value when dealing with post-perturbation instances, i.e., the difference between the solution value obtained by the solver before and after the perturbation. Moreover, the post-perturbation instance must ensure its optimal solution value does not deteriorate compared to the pre-perturbation state. 

In the future, we foresee a practical approach aimed at normalizing computation time across different hardware scenarios for various NCO solvers and conventional OR algorithms. This approach involves applying suitable scaling factors to facilitate fair comparisons. In summary, establishing well-accepted benchmark libraries and guidelines to compare the results of NCO solvers with OR algorithms is crucial. These resources would enhance researchers' understanding of the strengths and weaknesses of NCO solvers, thereby promoting their broader adoption by the OR community.

\section{Conclusion} 
\label{sec7}

This paper presents a systematic review to assess existing NCO solvers and achieves the following accomplishments while prior surveys did not, namely 1) all existing NCO solvers can be categorized accordingly w.r.t our proposed taxonomy; 2) the inadequacies of existing NCO solvers, including generalization, and the resolution of large-scale and multi-constrained VRPs, as well as on-going efforts to solve these inadequacies are comprehensively introduced; 3) the advantages of SL and UL learning paradigms are well presented; and 4) the difficulties in applying NCO solvers in real-world VRPs are discussed. Through a comprehensive investigation, we identify open inadequacies for future research. Notably, on-going efforts focus on only one or two of these inadequacies, with none attempting to address all of them. Furthermore, we propose numerous promising research directions to address these inadequacies. It must be acknowledged that due to the wide array of VRP variants, this survey solely delineates the inadequacies specific to the major variants. Nevertheless, this survey can alleviate the difficulties faced by fellow researchers dealing with non-traditional VRP variants and the other COPs. In addition, we provide a live web page repository for NCO solvers organized according to our proposed taxonomy, continuously updating with emerging NCO solvers. We anticipate that this survey and the live repository will achieve their intended objective of offering an overview of the latest developments in NCO solvers for VRPs, thereby serving as a valuable resource for researchers and practitioners.

At the same time, research on NCO solvers provides significant benefits to other scholars from related fields such as the broad machine learning and specific OR communities. For example, to reorder the Optical Character Recognition (OCR) text blocks into a logical reading sequence, Li et al. \cite{eccv} encoded each OCR text block as a node that encapsulates both visual layout and text semantic features, and subsequently employed a Ptr-Net-based network \cite{vinyals_pointer_2015} to perform the reordering. To generate a shorter advertisement video, Tang et al. \cite{ACCV} have also drawn on NCO solvers and proposed M-SAN, which first segments the original video and subsequently selects and assembles certain segments into a shorter video. Furthermore, investigating the geometric invariance properties of VRPs is also valuable for GNN-based point cloud analysis methods \cite{zhang2021revisiting}. Advancements in NCO solvers will surely benefit these relevant machine learning studies. In the OR community, several studies (e.g., \cite{xin_neurolkh_2021, zheng_combining_2021, ye_deepaco_2023}) have successfully integrated NCO techniques into traditional OR algorithms, improving their efficiency. For example, the classic LKH algorithm \cite{helsgaun_effective_2000} for solving VRP problems is computationally intensive because it requires a non-negligible number of iterations to obtain node penalties. Moreover, LKH relies on manually specified rules for constructing edge candidate sets, which could limit solution quality. To address this problem, Xin et al. \cite{xin_neurolkh_2021} proposed NeuroLKH to simultaneously predict the edge candidate sets and node penalties. Specifically, NeuroLKH begins by encoding edge and node inputs into feature vectors and then employs two distinct decoders to predict edge scores and node penalties, respectively. We anticipate that future research will further exploit the complementary strengths of OR and NCO to address different optimization problems.

\bibliography{survey}

\end{document}